\documentclass[10pt,journal,compsoc]{IEEEtran}

\usepackage{amssymb}
\usepackage{cite}
\usepackage{graphicx}
\usepackage{subfigure}
\usepackage{bbding}
\usepackage{multirow}
\usepackage{amsmath}
\usepackage{amssymb}
\usepackage{epstopdf}
\usepackage{url}
\usepackage{amsthm}

\usepackage{booktabs}
\usepackage{algorithm}
\usepackage{algorithmic}
\usepackage{hyperref}
\usepackage{chngpage}

\usepackage{multirow}
\usepackage[table,xcdraw]{xcolor}

\usepackage{fancyhdr}
\usepackage{amscd}
\usepackage[justification=centering]{caption}
\usepackage{booktabs}  
\usepackage{threeparttable}  
\usepackage{makecell}
\usepackage{graphicx}
\usepackage{array}
\usepackage{booktabs}

\theoremstyle{definition}
\newtheorem{theorem}{Definition}
\begin{document}
\title{Unsupervised Deep Anomaly Detection for \\Multi-Sensor Time-Series Signals}

\author{Yuxin~Zhang,
        Yiqiang~Chen,~\IEEEmembership{Senior Member,~IEEE,}
        Jindong~Wang,
        and~Zhiwen~Pan,~\IEEEmembership{Member,~IEEE}% <-this % stops a space

\IEEEcompsocitemizethanks{
\IEEEcompsocthanksitem Y. Zhang is with Global Energy Interconnection Development and Cooperation Organization, Xicheng District, Beijing, China, and Beijing Key Laboratory of Mobile Computing and Pervasive Device, Institute of Computing Technology, Chinese Academy of Sciences and University of Chinese Academy of Sciences, Beijing, China. E-mail: yuxin-zhang@geidco.org.
%\protect\\
%E-mail: yuxin-zhang@geidco.org.

\IEEEcompsocthanksitem Y. Chen and Z. Pan are with Beijing Key Laboratory of Mobile Computing and Pervasive Device, Institute of Computing Technology, Chinese Academy of Sciences and University of Chinese Academy of Sciences, Beijing, China. Y. Chen is also with Peng cheng Laboratory (PCL). E-mail: \{yqchen, pzw\}@ict.ac.cn (Corresponding author: Yiqiang Chen).
%\protect\\
%E-mail: \{yqchen, pzw\}@ict.ac.cn (Corresponding author: Yiqiang Chen).

\IEEEcompsocthanksitem J. Wang is with Microsoft Research Asia, Beijing, China, also correspondence. E-mail: Jindong.Wang@microsoft.com.
%\protect\\
%E-mail: Jindong.Wang@microsoft.com.

%\IEEEcompsocthanksitem Correspondence to: Y. Chen and J. Wang.
}
	
\thanks{Manuscript received April 19, 2005; revised August 26, 2015.}}

\markboth{}%
{Shell \MakeLowercase{\textit{et al.}}: Bare Demo of IEEEtran.cls for Computer Society Journals}



\IEEEtitleabstractindextext{%
\begin{abstract}
Nowadays, multi-sensor technologies are applied in many fields, e.g., Health Care (HC), Human Activity Recognition (HAR), and Industrial Control System (ICS). These sensors can generate a substantial amount of  multivariate time-series data. Unsupervised anomaly detection on multi-sensor time-series data has been proven critical in machine learning researches. The key challenge is to discover generalized normal patterns by capturing spatial-temporal correlation in multi-sensor data. Beyond this challenge, the noisy data is often intertwined with the training data, which is likely to mislead the model by making it hard to distinguish between the normal, abnormal, and noisy data. Few of previous researches can jointly address these two challenges. In this paper, we propose a novel deep learning-based anomaly detection algorithm called Deep Convolutional Autoencoding Memory network (CAE-M). We first build a Deep Convolutional Autoencoder to characterize spatial dependence of multi-sensor data with a Maximum Mean Discrepancy (MMD) to better distinguish between the noisy, normal, and abnormal data. Then, we construct a Memory Network consisting of linear (Autoregressive Model) and non-linear predictions (Bidirectional LSTM with Attention) to capture temporal dependence from time-series data. Finally, CAE-M jointly optimizes these two subnetworks. We empirically compare the proposed approach with several state-of-the-art anomaly detection methods on HAR and HC datasets. Experimental results demonstrate that our proposed model outperforms these existing methods.
\end{abstract}

\begin{IEEEkeywords}
Unsupervised anomaly detection, Multi-sensor time series, Convolutional autoencoder, Attention based BiLSTM.
\end{IEEEkeywords}}

\maketitle

\IEEEdisplaynontitleabstractindextext

% \fi
\IEEEpeerreviewmaketitle

\IEEEraisesectionheading{\section{Introduction}}
\label{sec:introduction}

\IEEEPARstart{A}{nomaly} detection has been one of the core research areas in machine learning for decades, with wide applications such as cyber-intrusion detection \cite{chalapathy2019deep}, medical care \cite{zhang2018deep}, sensor networks \cite{ball2017comprehensive}, video anomaly detection \cite{kiran2018overview} and so on. 
Anomaly detection seems to be a simple two-category classification, i.e., we can learn to classify the normal or abnormal data. However, it is also faced with the following challenges. First, training data is highly imbalanced since the anomalies are often extremely rare in a dataset compared to the normal instances. Standard classifiers try to maximize accuracy in classification, so it often falls into the trap of overlapping problem, which means that the model classifies the overlapping region as belonging to the majority class while assuming the minority class as noise. Second, there is no easy way for users to manually label each training data, especially the anomalies. In many cases, it is prohibitively hard to represent all types of anomalous behaviors. Due to above challenges, there is a growing trend to use unsupervised learning approaches for anomaly detection compared with semi-supervised and supervised learning approaches since unsupervised methods can handle the imbalanced and unlabeled data in a more principled way \cite{ramchandran2018unsupervised,pang2021deep,2020A,zhang2021deep,chen2019cross}. 

Nowadays, the prevalence of sensors in machine learning and pervasive computing research areas such as Health Care (HC)~\cite{chen2020fedhealth,wang2018deep} and Human Activity Recognition (HAR)~\cite{wang2019deep,wang2018stratified} generate a substantial amount of multivariate time-series data. These learning algorithms based on multi-sensor time-series signals give priority to dealing with spatial-temporal correlation of multi-sensor data. Many approaches for spatial-temporal dependency amongst multiple sensors \cite{li2019mad,zhang2019deep,patraucean2015spatio} have been studied. It seems intuitive to apply previous unsupervised anomaly detection methods on multi-sensor time-series data. Unfortunately, there are still several challenges. 

First, anomaly detection in spatial-temporal domain becomes more complicated due to the temporal component in time-series data. Conventional anomaly detection techniques such as PCA \cite{paffenroth2013space}, k-means \cite{latecki2007outlier}, OCSVM \cite{scholkopf2001estimating} and Autoencoder \cite{sakurada2014anomaly} are unable to deal with multivariate time-series signals since they cannot simultaneously capture the spatial and temporal dependencies. 
% To address this issue, our model design is to discover generalized pattern of normal data by capturing spatial-temporal correlation in high-dimensional data;
Second, these reconstruction-based models such as Convolutional AutoEncoders (CAEs) \cite{hasan2016learning} and Denoising AutoEncoders (DAEs) \cite{vincent2008extracting} are usually used for anomaly detection. It is generally assumed that the compression of anomalous samples is different from that on normal samples, and the reconstruction error becomes higher for these anomalous samples. In reality, being influenced by the high complexity of model and the noise of data, the reconstruction error for the abnormal input could also be fit so well by the training model \cite{zong2018deep,gong2019memorizing}. That is, the model is robust to noise and anomalies. 
% Hence, we need to observe the changes in low-dimensional features and the changes of distribution over the samples in a more granular way, thus distinguishing between normal and abnormal data obviously.
Third, in order to reduce the dimensionality of multi-sensor data and detect anomalies, two-step approaches are widely adopted. As for the drawback of some works \cite{li2018anomaly, aliakbarisani2019data}, the joint performance of two baseline models can easily get stuck in local optima, since two models are trained separately. 
% Therefore, we propose an end-to-end hybrid model by simultaneously minimizing compound objective function to obtain global optima.

In order to solve the above three challenges, this paper presents a novel unsupervised deep learning based anomaly detection approach for multi-sensor time-series data called Deep Convolutional Autoencoding Memory network (CAE-M). The CAE-M network composes of two main sub-networks: characterization network and memory network. Specifically, we employ deep convolutional autoencoder as feature extraction module, with attention-based Bidirectional LSTMs and Autoregressive model as forecasting module. By simultaneously minimizing reconstruction error and prediction error, the CAE-M model can be jointly optimized. During the training phase, the CAE-M model is trained to explicitly describe the normal pattern of multi-sensor time-series data. During the detection phase, the CAE-M model calculate the compound objective function for each captured testing data. Through combining these errors as a composite anomaly score, a fine-grained anomaly detection decision can be made. To summarize, the main contributions of this paper are four-fold:

1) The proposed composite model is designed to characterize complex spatial-temporal patterns by concurrently performing the reconstruction and prediction analysis. In reconstruction analysis, we build Deep Convolutional Autoencoder to fuse and extract low-dimensional spatial features from multi-sensor signals. In prediction analysis, we build Attention-based Bidirectional LSTM to capture complex temporal dependencies. Moreover, we incorporate Auto-regressive linear model in parallel to improve the robust and adapt for different use cases and domains.
%And the corresponding anomaly score based on compound objective function is formated for fine-grained anomaly detection.

2) To reduce the influence of noisy data, we improve Deep Convolutional Autoencoder with a Maximum Mean Discrepancy (MMD) penalty. MMD is used to encourage the distribution of the low-dimensional representation to approximate some target distribution. It aims to make the distribution of noisy data close to the distribution of normal training data, thereby reducing the risk of overfitting. Experiments demonstrate that it is effective to enhance the robustness and generalization ability of our method. 
%The aim is to discover generalized pattern of normal data by reducing the error between input sequence and reconstructed sequence.

3) The CAE-M is an end-to-end learning model that two sub-networks can co-optimize by a compound objective function with weight coefficients. This single-stage approach can not only streamline the learning procedure for anomaly detection, but also avoid the model getting stuck in local minimum through joint optimization.
%the parameters rather than optimizing them separately.
%joint optimization

4) Experiments on three multi-sensor time-series datasets demonstrate that CAE-M model has superior performance over state-of-the-art techniques. In order to further verify the effect of our proposed model, fine-grained analysis, effectiveness evaluation, parameter sensitivity analysis and convergence analysis show that all the components of CAE-M together leads to the robust performance on all datasets.

The rest of the paper is organized as follows. Section \ref{sec:related} provides an overview of existing methods for anomaly detection. Our proposed methodology and detailed framework is described in Section \ref{sec:method}. Performance evaluation and analysis of experiment is followed in Section \ref{sec:exp}. Finally, Section \ref{sec:con} concludes the paper and sketches directions for possible future work.

\section{Related Work}
\label{sec:related}
Anomaly detection has been studied for decades. Based on whether the labels are used in the training process, they are grouped into supervised, semi-supervised and unsupervised anomaly detection. Our main focus is the unsupervised setting. In this section, we demonstrate various types of existing approaches for unsupervised anomaly detection, which can be categorized into traditional anomaly detection and deep anomaly detection.

\subsection{Traditional anomaly detection}
Conventional methods can be divided into three categories. 1) Reconstruction-based methods are proposed to represent and reconstruct accurately normal data by a model, for example, PCA \cite{paffenroth2013space}, Kernel PCA \cite{scholkopf1998nonlinear,hoffmann2007kernel} and Robust PCA \cite{paffenroth2018robust}. Specifically, RPCA is used to identify a low rank representation including random noise and outliers by using a convex relaxation of the rank operator; 2) Clustering analysis is used for anomaly detection, such as Gaussian Mixture Models (GMM) \cite{laxhammar2009anomaly}, k-means \cite{latecki2007outlier} and Kernel Density Estimator (KDE) \cite{kim2012robust}. They cluster different data samples and find anomalies via a predefined outlierness score; 3) the methods of one-class learning model are also widely used for anomaly detection. For instance, One-Class Support Vector Machine (OCSVM) \cite{scholkopf2001estimating} and Support Vector Data Description (SVDD) \cite{tax2004support} seek to learn a discriminative hypersphere surrounding the normal samples and then classify new data as normal or abnormal.

It is notable that these conventional methods for anomaly detection are designed for static data. To capture the temporal dependencies appropriately, Autoregression (AR) \cite{gunnemann2014robust}, Autoregressive Moving Average (ARMA) \cite{hamilton1994time} and Autoregressive Integrated Moving Average (ARIMA) model \cite{moayedi2008arima} are widely used. These models represent time series that are generated by passing the input through a linear or nonlinear filter which produces the output at any time using the previous output values. Once we have the forecast, we can use it to detect anomalies and compare with groundtruth. Nevertheless, AR model and its variants are rarely used in multi-sensor multivariate time series due to their high computational cost.
%GAN 这种生成式模型就不提了在这里
\subsection{Deep anomaly detection}
In deep learning-based anomaly detection, the reconstruction models, forecasting models as well as composite models will be discussed.
\subsubsection{Reconstruction models}
The reconstruction model focuses on reducing the expected reconstruction error by different methods. For instance, Autoencoders \cite{sakurada2014anomaly} are often utilized for anomaly detection by learning to reconstruct a given input. The model is trained exclusively on normal data. Once it is not able to reconstruct the input with equal quality compared to the reconstruction of normal data, the input sequence is treated as anomalous data. LSTM Encoder-Decoder model \cite{malhotra2016lstm} is proposed to learn temporal representation of the input time series by LSTM networks and use reconstruction error to detect anomalies. Despite its effectiveness, LSTM does not take spatial correlation into consideration. Convolutional Autoencoders (CAEs) \cite{hasan2016learning} are an important method of video anomaly detection, which are able of capturing the 2D image structure since the weights are shared among all locations in the input image. Furthermore, since Convolutional long short-term memory (ConvLSTM) can model spatial-temporal correlations by using convolutional layers instead of fully connected layers, some researchers \cite{zhang2019deep,luo2017remembering} add ConvLSTM layers to autoencoder, which better encodes the change of appearance for normal data.

Variational Autoenocders (VAEs) are a special form of autoencoder that models the relationship between two random variables, latent variable $z$ and visible variable $x$. A prior for $z$ is usually multivariate unit Gaussian $\mathcal N(0,I)$. For anomaly detection, authors \cite{an2015variational} define the reconstruction probability that is the average probability of the original data generating from the distribution. Data points with high reconstruction probability is classified as anomalies, vice versa. Others like Denoising AutoEncoders (DAEs) \cite{vincent2008extracting}, Deep Belief Networks (DBNs) \cite{wulsin2010semi} and Robust Deep Autoencoder (RDA) \cite{zhou2017anomaly} have also been reported good performance for anomaly detection.

\subsubsection{Forecasting models}
The forecasting model can also be used for anomaly detection. It aims to predict one or more continuous values, e.g. forecasting the current output values $x_t$ for the past $p$ values $[x_{t-p},...,x_{t-2},x_{t-1}]$. RNN and LSTM is the standard model for sequence prediction. In the work \cite{filonov2017rnn,ergen2017unsupervised}, authors perform anomaly detection by using RNN-based forecasting models to predict values for the next time period and minimize the mean squared error (MSE) between predicted and future values. Recently, there have also been attempted to perform anomaly detection using other feed-forward networks. For instance, Shalyga \textit{et al.} \cite{shalyga2018anomaly} develop Neural Network (NN) based forecasting approach to early anomaly detection. Kravchik and Shabtai \cite{kravchik2018detecting} apply different variants of convolutional and recurrent networks to perform forecasting model. And the results show that 1D convolutional networks obtain the best accuracy for anomaly detection in industrial control systems. In another work \cite{lai2018modeling}, Lai \textit{et al.} propose a forecasting model, which uses CNN and RNN, namely LSTNet, to extract short-term local dependency pattern and long-term pattern for multivariate time series, and incorporates Linear SVR model in the LSTNet model. Besides, other efforts have been performed in \cite{liu2018future} using GAN-based anomaly detection. The model adopts U-Net as generator to predict next frame in video and leverages the adversarial training to discriminate whether the prediction is real or fake, thus abnormal events can be easily identified by comparing the prediction and ground truth.

\subsubsection{Composite models}
Besides single model, composite model for unsupervised anomaly detection has gained a lot attention recently. Zong \textit{et al.} \cite{zong2018deep} utilize a deep autoencoder to generate a low-dimensional representation and reconstruction error, which is further fed into a Gaussian Mixture Model to model density distribution of multi-dimensional feature. However, they cannot consider the spatial-temporal dependency for multivariate time series data. Different from this work, the Composite LSTM model \cite{srivastava2015unsupervised} uses single encoder LSTM and multiple decoder LSTMs to perform different tasks such as reconstructing the input sequence and predicting the future sequence. In \cite{medel2016anomaly}, the authors use ConvLSTM model as a unit within the composite LSTM model following a branch for reconstruction and another for prediction. This type of composite model is currently used to extract features from video data for the tasks of action recognition. Similarly, authors in \cite{zhao2017spatio} propose Spatial-Temporal AutoEncoder (STAE) for video anomaly detection, which utilizes 3D convolutional architecture to capture the spatial-temporal changes. The architecture of the network is an encoder followed by two branches of decoder for reconstructing past sequence and predicting future sequence respectively.

As mentioned above, unsupervised anomaly detection techniques have still many deficiencies. For traditional anomaly detection, it is hard to learn representations of spatial-temporal patterns in multi-sensor time-series signals. For a reconstruction model, a single task could make the model suffer from the tendency to store information only about the inputs that are memorized by the AE. And for the forecasting model, this task could suffer from only storing the last few values that are most important for predicting the future \cite{srivastava2015unsupervised,medel2016anomaly}. Hence, their performance will be limited since model only learn trivial representations. For composite model, these researchers design their models for different purposes. Zong \textit{et al.} \cite{zong2018deep} could solve problem that the model is robust to noise and anomalies through performing density estimation in a low-dimensional space. Zhao \textit{et al.} \cite{zhao2017spatio} could consider the spatial-temporal dependency through 3D convolutional reconstructing and forecasting architectures. However, few studies could address these issues simultaneously.

Different from these works, our research makes the following contributions: 1) The proposed model is designed to characterize complex spatial-temporal dependencies, thus discovering generalized pattern of multi-sensor data; 2) Adding a Maximum Mean Discrepancy (MMD) penalty could avoid the model generalizing so well for noisy data and anomalies; 3) Combining Attention-based Bidirectional LSTM (BiLSTM) and traditional Auto-regressive linear model could boost the model's performance from different time scale; 4) The composite baseline model is generated based on end-to-end training which means all the components within the model are jointly trained with compound objective function.

Besides, some learning algorithms based on time-series data have been studied for decades. \cite{zhang2018salient} propose Unsupervised Salient Subsequence Learning to extract subsequence as new representations of the time series data. Due to the internally sequential relationship, many neural network-based models can be applied to time series in an unsupervised learning manner. For example, some 1D-CNN models \cite{fawaz2020inceptiontime,tang2020rethinking} have been proposed to solve time series tasks with a very simple structure and the sota performance. Moreover, the multiple time series signal usually has some kinds of co-relations, \cite{wu2020connecting} propose a method to learn the relation graph on multiple time series. Some anomaly detection based on multiple time series applications are available for wastewater treatment \cite{jun2011fault}, for ICU \cite{chen2018dynamic}, and for sensors \cite{zhang2018multi}.

\begin{figure*}[t!]
	\centering
	\includegraphics[width=1.\linewidth]{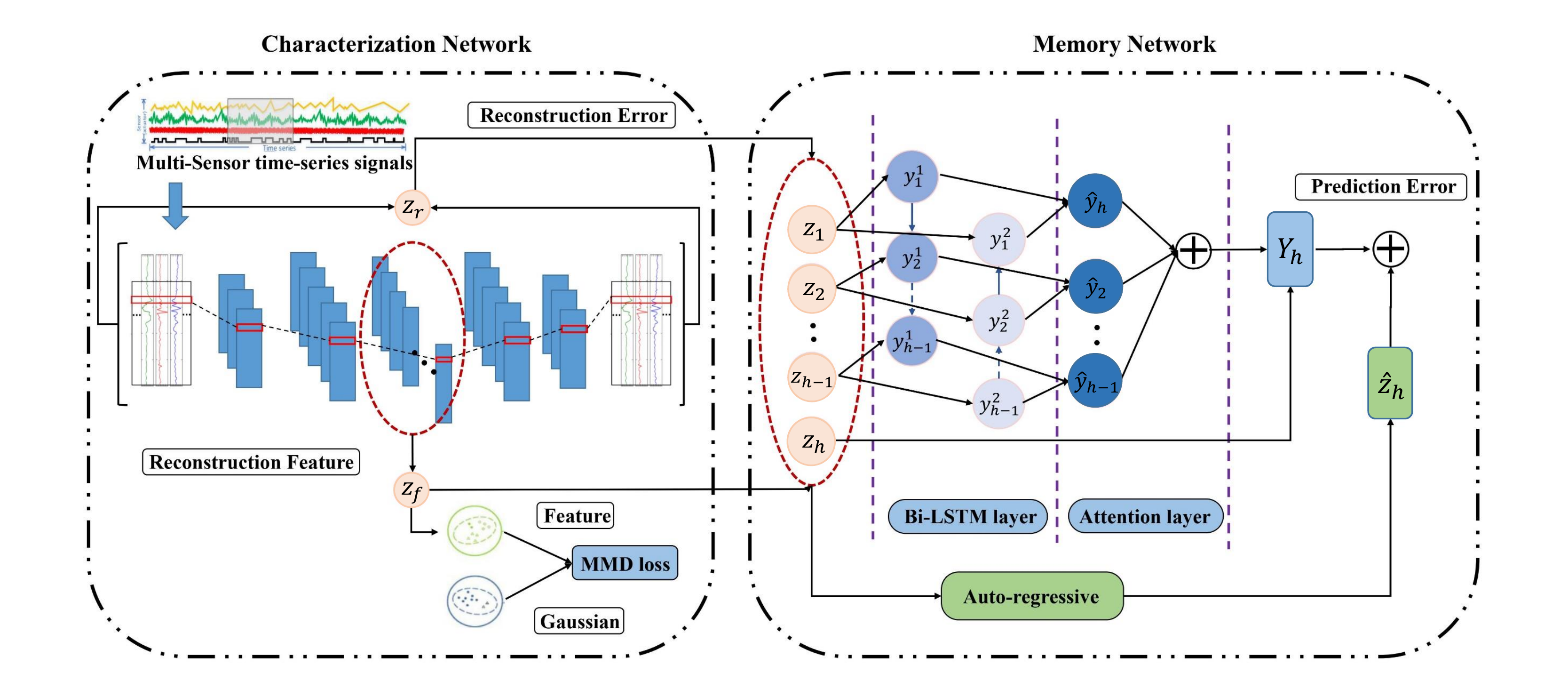}
	\caption{The overview of the proposed CAE-M model.}
	\label{fig1}
	\vspace{-.2in}
	\label{overview}
\end{figure*}

\section{The Proposed Method}
\label{sec:method}
\subsection{Notation} %问题定义

In a multi-sensor time series anomaly detection problem, we are given a dataset generated by $n$ sensors ($n > 1$). Without loss of generality, we assume each sensor generates $m$ signals (e.g., an accelerometer often generates 3-axis signals). Denote $\mathcal{S}$ the signal set, we have $N=|\mathcal{S}|=nm$ signals in total. For each signal $x_i \in \mathcal{S}$, $x_i \in \mathbb{R}^{t_i \times 1}$, where $t_i$ denotes the length of signal $x_i$. Note that even each sensor signal may have different length, we are often interested in their intersections, i.e., all sensors are having the same length $T$, i.e., $X = (x_1,\cdots,x_N)^T \in \mathbb{R}^{N \times T}$ denotes an input sample containing all sensors.

\begin{theorem}[Unsupervised anomaly detection]

It is non-trivial to formally define an anomaly. In this paper, we are interested in detecting anomalies in a classification problem. Let $\mathcal{Y}=\{1, 2, \cdots, K\}$ be the classification label set, and $K$ the total number of classes, then the dataset $\mathcal{D} = (X_i, y_i)^N_{i=1}$. Eventually, our goal is to detect whether an input sample $X_a$ belongs to one of the $K$ predefined classes with a high confidence. If not, then we call $X_a$ an anomaly. Note that in this paper, we are dealing with an unsupervised anomaly detection problem, where the labels are unseen during training, which is more obviously challenging.

\end{theorem}

\subsection{Overview} %方法分为几个部分，分别解决了什么挑战

There are some existing works \cite{scholkopf2001estimating,hasan2016learning,li2018anomaly} attempting to resolve the unsupervised anomaly detection problem. Unfortunately, they may face several critical challenges. First, conventional anomaly detection techniques such as PCA \cite{paffenroth2013space}, k-means \cite{latecki2007outlier} and OCSVM \cite{scholkopf2001estimating} are unable to capture the temporal dependencies appropriately because they cannot deliver temporal memory states. Second, since the normal samples might contain noise and anomalies, using deep anomaly detection approaches such as standard Autoencoders \cite{hasan2016learning,sakurada2014anomaly} is likely to affect the generalization capability. Third, the multi-stage approaches, i.e., feature extraction and predictive model building are separated \cite{li2018anomaly, aliakbarisani2019data}, can easily get stuck in local optima.

In this paper, we present a novel approach called Convolutional Autoencoding Memory network (CAE-M) to tackle the above challenges. \figurename~\ref{overview} gives an overview of the proposed method. In a nutshell, CAE-M is built upon a convolutional autoencoder, which is then fed into a predictive network. Concretely, we encode the spatial information in multi-sensor time-series signals into the low-dimensional representation via Deep Convolutional Autoencoder (CAE). In order to reduce the effect of noisy data, some existing works have tried to add Memory module \cite{gong2019memorizing} or Gaussian Mixture Model (GMM) \cite{zong2018deep}. In our proposed method, we simplify these modules into penalty item, which called Maximum Mean Discrepancy (MMD) penalty. Adding a MMD term can encourage the distribution of training data to approximate the same distribution such as Gaussian distribution, thus reducing the risk of overfitting caused by noise and anomalies in training data \cite{zong2018deep}. And then we feed the representation and reconstruction error to the subsequent prediction network based on Bidirectional LSTM (Bi-LSTM) with Attention mechanism and Auto-regressive model (AR) which could predict future feature values by modeling the temporal information. Through the composite model, the spatial-temporal dependencies of multi-sensor time-series signals can be captured. Finally, we propose a compound objective function with weight coefficients to guide end-to-end training. For normal data, the reconstructed value generated by data coding is similar to the original input sequence and the predicted value is similar to the future value of time series, while the reconstructed value and the predicted value generated by abnormal data change greatly. Therefore, in inference process, we can detect anomalies precisely by computing the loss function in composite model.

%mmd看的是在整体样本（时间序列窗口内或一个batch内）的分布是否相同
\subsection{Characterization Network}
In the characterization network, we perform representative learning by fusing multivariate signals in multiple sensors. The low-dimensional representation contains two components: (1) the features which are abstracted from the multivariate signals; (2) the reconstruction error over the distance metrics such as Euclidean distance and Minkowski distance. To avoid the autoencoder generalizing so well for abnormal inputs, optimization function combines reconstruction loss by measuring how close the reconstructed input is to the original input and the regularization term by measuring the similarity between the two distributions (i.e., the distribution of low-dimensional features and Gaussian distribution).

\subsubsection{Deep feature extraction}
We employ a deep convolutional autoencoder to learn the low-dimensional features. Specifically, given $N$ time series with length $T$, we pack into a matrix $x \in \mathbb{R}^{N \times T} $ with multi-sensor time-series data. The matrix is then fed to deep convolutional autoencoder (CAE). The CAE model is composed of two parts, an encoder and a decoder as in Eq.~\eqref{z1} and Eq.~\eqref{z2}. Assuming that $x'$ denotes the reconstruction of the same shape as $x$, the model is to compute low-dimensional representation $z_f$, as follows:
\begin{equation}
\label{z1}
\begin{aligned}
& z_f = Encode(x), 
\end{aligned}
\end{equation}
\begin{equation}
\label{z2}
\begin{aligned}
& x' = Decode(z_f).
\end{aligned}
\end{equation}

The encoder in Eq.~\eqref{z1} maps an input matrix $x$ to a hidden representation $z_f$ by many convolutional and pooling layers. Each convolutional layer will be followed by a max-pooling layer to reduce the dimensions of the layers. A max-pooling layer pools features by taking the maximum value for each patch of the feature maps and produce the output feature map with reduced size according to the size of pooling kernel.

The decoder in Eq.~\eqref{z2} maps the hidden representation $z_f$ back to the original input space as a reconstruction. In particular, a decoding operation needs to convert from a narrow representation to a wide reconstructed matrix, therefore the transposed convolution layers are used to increase the width and height of the layers. They work almost exactly the same as convolutional layers, but in reverse. 

The difference between the original input vector $x$ and the reconstruction $x'$ is called the reconstruction error $z_r$. The error typically used in the autoencoder is Mean Squared Error (MSE), which measures how close the reconstructed input $x'$ is to the original input $x$, as follows in Eq.~\eqref{eq-mse}. 

\begin{equation}
\label{eq-mse}
L_{MSE} = \lVert x - x^ \prime \lVert ^2_2,
\end{equation}
where $\lVert \cdot \lVert^2_2$ is the $\textit{l}_2$-norm.

\subsubsection{Handling noisy data}
To reduce the influence of noisy data, we need to observe the changes in low-dimensional features and the changes of distribution over the samples in a more granular way, thus distinguishing between normal and abnormal data obviously.

%The loss function aims for optimizing model. 
Inspired by \cite{zong2018deep}, in order to avoid the autoencoder generalizing so well for noisy data and abnormal data, we hope to detect "lurking" anomalies that reside in low-density areas in the reduced low-dimensional space. Our proposed method is conceptually similar to Gaussian Mixture Model (GMM) as target distributions. The loss function is complemented by MMD as a regularization term that encourages the distribution of the low-dimensional representation to be similar to a target distribution. It aims to make the distribution of noisy data close to the distribution of normal training data, thereby reducing the risk of overfitting. Specifically, Maximum Mean Discrepancy (MMD) \cite{smola2007hilbert} is a distance-measure between the samples of the distributions. Given the latent representation $z_a=\{z_f^{(1)},...,z_f^{(h)}\} \in \mathbb{R}^{h \times d}$, where $d$ is a latent space (usually $d < N \times T$) and $h$ denotes all of the time steps at one iteration. For CAE with MMD penalty, the Gaussian distribution $P_z$ in reproduction kernel Hilbert space $\mathcal{H}$ is chosen as the target distribution. We compute the Kernel MMD as the follows:
\begin{equation}
\label{z7}
\begin{aligned}
L_{MMD}(Z,P_z) = ||\frac 1 h \sum_{i=1}^{h} \phi (z_f^{(i)})- \frac 1 h \sum_{i=1}^{h} \phi (z^{(i)})|| ^2_\mathcal{H}.
\end{aligned}
\end{equation}
Here we have the distribution $Z$ of the low-dimensional representation $z_f^{(i)}$ and the target distribution $z^{(i)} \sim P_z$ over a set $\mathcal{X}$. The MMD is defined by a feature map $\phi : \mathcal{X} \rightarrow \mathcal{H}$ where $\mathcal{H}$ is a reproducing kernel Hilbert space (RKHS).

During the training process, we could apply the kernel trick to compute the MMD. And it turns out that many kernels, including the Gaussian kernel, lead to the MMD being zero if and only the distributions are identical. Letting $k(x,y)=\left \langle \phi (x), \phi (y)\right \rangle _ \mathcal{H}$, we yield an alternative characterization of the MMD as follows:

\begin{equation}
\label{z8}
\begin{aligned}
L_{MMD}(Z,P_z) = &||\frac 1 {h^2} \sum_{i \neq j} k(z_f^{(i)},z_f^{(j)})+\frac 1 {h^2} \sum_{i \neq j} k(z^{(i)},z^{(j)})\\
&-\frac 2 {h^2} \sum_{i , j} k(z_f^{(i)},z^{(j)})|| _\mathcal{H}.
\end{aligned}
\end{equation}
Here the kernel is defined as $k(u,v)=exp(- \frac {||u-v||^2} {2 \sigma ^2})$. The latent representation with Gaussian distribution $P_z$ is performed by sampling from $P_z$ and approximating by averaging the kernel $k(\cdot,\cdot)$ evaluated at all pairs of samples.

Note that we usually do batch training for neural network training. It means that the model is trained using a subsample of data at one iteration. In this work, we need to compute the MMD over a set of $\mathcal{X}$ at one iteration, where the number of $\mathcal{X}$ is equal to $batch size \times time step$. That is, the latent representation is denoted as $z_a=\{z_f^{(1)},...,z_f^{(l)}\} \in \mathbb{R}^{l \times d}$, where $l=batch size \times h$.
\subsection{Memory Network}

To simultaneously capture the spatial and temporal dependencies, our proposed model is designed to characterize complex spatial-temporal patterns by concurrently performing the reconstruction analysis and prediction analysis. Considering the importance of temporal component in time series, we propose non-linear prediction and linear prediction to detect anomalies by comparing the future prediction and the next value appearance in the feature space.

The characterization network generates feature representations, which include reconstruction error and reduced low-dimensional features learned by the CAE at $h$ time steps. Denote input features as $z_h$ for $h=1,...,H$:
\begin{equation}
\label{z9}
\begin{aligned}
z_h = [z_f,z_r]_h, \ h \in [1,H].
\end{aligned}
\end{equation}

Our goal is to predict the current value $z_h$ for the past values $[z_{1},z_2,...,z_{h-1}]$. The memory network combines non-linear function  based predictor and linear function based predictor to tackle temporal dependency problem.

\subsubsection{Non-linear prediction}

Non-linear predictor function has different types such as Recurrent neural networks (RNNs), Long Short-Term Memory (LSTM)\cite{greff2017lstm} and Gated Recurrent Unit (GRU) \cite{chung2014empirical}. Original RNNs fall short of learning long-term dependencies. In this work, we adopt a Bidirectional LSTM with attention mechanism \cite{liu2019bidirectional} which could consider the whole/local context while calculating the relevant hidden states. Specifically, the Bidirectional LSTM (BiLSTM) runs the input in two ways, one LSTM from past to future and one LSTM from future to past. Different from unidirectional, the two hidden states combined are able in any point in time to preserve information from both past and future. A BiLSTM unit consists of four components: input gate $i_h$, forget gate $f_h$, output gate $o_h$ and cell activation vector $c_h$. The hidden state $y_h$ given input $z_h$ is computed as follows:
%\begin{subequations}
\begin{equation}
\label{z10}
i_h = \sigma (W_{zi}z_h + W_{yi} y_{h-1} + W_{ci} c_{h-1} + b_i),
\end{equation}
\begin{equation}
\label{z11}
f_h = \sigma (W_{zf}z_h + W_{yf} y_{h-1} + W_{cf} c_{h-1} + b_f), 
\end{equation}
\begin{equation}
\label{z12}
o_h = \sigma (W_{zo}z_h + W_{yo} y_{h-1} + W_{co} c_{h-1} + b_o), 
\end{equation}
\begin{equation}
\label{z13}
\widetilde{c_h} = tanh(W_{zc}z_h + W_{yc} h_{h-1} + W_{cc} c_{h-1} + b_c),
\end{equation}
\begin{equation}
\label{z14}
c_h = f_h \otimes c_{h-1} + i_h \otimes \widetilde{c_h},
\end{equation}
\begin{equation}
\label{z15}
y_h = o_h \otimes tanh(c_h),
\end{equation}
\begin{equation}
\label{z16}
%y_h = softmax (W_{yy} \ast [\stackrel{\rightarrow}{y_h} \oplus \stackrel{\leftarrow}{y_h}] )
\hat{y}_h = [y^1_h ; y^2_h ],
%A \xleftarrow{n=0} B \xrightarrow[T]{n>0} C
\end{equation}
%\end{subequations}
where $i_h$, $f_h$, $o_h$, $c_h$ represent the value of $i,f,o,c$ at the moment $h$ respectively, $W$ and $b$ denote the weight matrix and bias vector, $\sigma (\cdot)$ and $tanh (\cdot)$ are activation function, the operator $\otimes$ denotes element-wise multiplication, the current cell state $c_h$ consists of two components, namely previous memory $c_{h-1}$ and modulated new memory $\widetilde{c_h}$, the output $\hat{y}_h$ combines the forward $y^1_h$ and backward $y^2_h$ pass outputs. Note that the merge mode by which outputs of the forward and backward are combined has different types, e.g. sum, multiply, concatenate, average. In this work, we use the mode ``sum'' to obtain the output $\hat{y}_h$.

Attention mechanism for processing sequential data that could focus on the features of the keywords to reduce the impact of non-key temporal context. Hence, we adopt temporal attention mechanism to produce a weight vector and merge raw features from each time step into a segment-level feature vector, by multiplying the weight vector. The work process of attention mechanism is following detailed.
\begin{equation}
M_h = tanh(W_{h}\hat{y}_h + b_{h}),
\end{equation}
\begin{equation}
E_h = \sigma (W_a M_h + b_a), 
\end{equation}
\begin{equation}
A_h = softmax(E_h),
\end{equation}
\begin{equation}
Y_h = \sum_{h} A_h \ast \hat{y}_h.
\end{equation}
Here $W$ and $b$ are represented as the weight and bias. A weighted sum of the $\hat{y}_h$ based on the weight $A_h$ is computed as the context representation $Y_h$. The context representation is considered as the predicted value of $z_h$ for temporal features $[z_1,z_2,...,z_{h-1}]$.

%Similar to $y^1_h$, $y^2_h$ could be calculated into the backward context representation $F_h$. Finally, the output $S = [L_h,F_h]$  is obtained. 

\subsubsection{Linear prediction}
%https://towardsdatascience.com/arima-sarima-vs-lstm-with-ensemble-learning-insights-for-time-series-data-509a5d87f20a
%https://eigenfoo.xyz/deep-autoregressive-models/
Autoregressive (AR) model is a regression model that uses the dependencies between an observation and a number if lagged observations. Non-linear Recurrent Networks are theoretically more expressive and powerful than AR models. In fact, AR models also yield good results in forecasting short term modeling. In specific real datasets, such infinite-horizon memory isn't always effective. Therefore, we incorporate AR model in parallel to the non-linear memory network part.

The AR model is formulated as follows:
\begin{equation}
\label{z17}
\begin{aligned}
\hat{z}_h = c \sum_{i=1}^{h-1} w_{h-i} + \sum_{i=1}^{h-1} w_{h-i}*z_{h-i}, 
\end{aligned}
\end{equation}
where $w_1,...,w_{h-1}$ are the weights of the AR model, $c$ is a constant, $\hat{z}_h$ represents the predicted value for past temporal value $[z_1,z_2,...,z_{h-1}]$. We implement this model using Dense layer of network to combine the weights and data.

In the output layer, the prediction error is obtained by computing the difference between the output of predictor model and true value ${z_h}$. The final prediction error integrates the output of non-linear prediction model and linear prediction model. The following equation is written as:
\begin{equation}
\label{z18}
\begin{aligned}
L_{predict} = \sum_{h \in \Omega _{batch}} (\underbrace{||Y_{h} - z_{h}||^2_F}_{\text{Attention-based BiLSTM}} + \underbrace{||\hat{z}_{h} - z_{h}||^2_F }_{\text{Autoregressive}}),
\end{aligned}
\end{equation}
where $\Omega _{batch}$ is a subsample of training data, $||\cdot||_F$ is the Frobenius norm.

\subsection{Joint optimization}
As for multi-step approach, it can easily get stuck in local optima, since models are trained separately. Therefore, we propose an end-to-end hybrid model by minimizing compound objective function.

The CAE-M objective has four components, MSE (reconstruction error) term, MMD (regularization) term, prediction error (non-linear forecasting task) term and prediction error (linear forecasting task) term. Given $X$ samples $\{x_1,x_2,...,x_D\}, x_i \in \mathbb{R}^{N \times T}$, the objective function is constructed as:
\begin{equation}
\label{z19}
\begin{split}
J(\theta) & = L_{MSE} + \lambda _1 \cdot L_{MMD} + \lambda _2 \cdot L_{lp} + \lambda _3 \cdot L_{np} \\
& = \frac {1} {M} \sum_{i=1}^{M} L(x_i,x^\prime_i) + \lambda_1 L_{MMD}(Z,P_Z)  \\
& + \frac 1 M \sum_{i=1}^{M} [\lambda_2 ||Y_{h}^{(i)} - z_{h}^{(i)}||^2_F + \lambda_3 ||\hat{z}_{h}^{(i)} - z_{h}^{(i)}||^2_F],  %\Omega (z_f)
\end{split}
\end{equation}
where $M$ is batch size used for training, $h$ is current time step, $\lambda_1$,$\lambda_2$ and $\lambda_3$ are the meta parameters controlling the importance of the loss function.

Restating our goals more formally, we would like to:
\begin{itemize}
	\item Minimize the reconstruction error in the characterization network, that is, minimize the error in reconstructing $x'$ from $x$ at all time step $h$. We need to compute the average error at each time step of sample. The purpose is to obtain better low-dimensional representation for multi-sensor data.
\end{itemize}
\begin{itemize}
	\item Minimize the MMD loss that encourages the distribution $Z$ of the low-dimensional representation to be similar to a target distribution $P_z$. It can make anomalies deviate from normal data in the reduced dimensions.
\end{itemize}
%\begin{itemize}
%https://www.deeplearningbook.org/contents/autoencoders.html
%https://www.quora.com/What-is-the-purpose-of-sparsity-constraint-in-autoencoder
%	\item Add sparsity constraints in the last encoder layer. The constraint forces the CAE to represent each input using only a small number of hidden neurons. Different inputs trigger different sets of neurons in the hidden layer, and only the ones capable of responding to meaningful patterns in the input data will be fired at appropriate instances.
%\end{itemize}
\begin{itemize}
	%https://www.deeplearningbook.org/contents/autoencoders.html
	\item Minimize the prediction error by integrating non-linear predictor and linear predictor. We split the set $\{z_1,z_2,...,z_h\}$ obtained by characterization network into the current value $z_h$ and the past values $[z_1,z_2,...,z_{h-1}]$. And then the predicted values $Y_{h}$ and $\hat{z}_h$ are obtained by minimizing prediction errors. The purpose is to accurately express the information of the next temporal slice using different predictor, thus updating low-dimensional feature and reconstruction error.
\end{itemize}
\begin{itemize}
	\item $\lambda_1$, $\lambda_2$ and $\lambda_3$ are the meta parameters in CAE-M. In practice, $\lambda_1= e-04$, $\lambda_2= 0.5$, and $\lambda_3= 0.5$ usually achieve desirable results. Here MMD is complemented as a regularization term. The parameter selection is performed in Section \ref{sec:parameter}.
\end{itemize}

\begin{table*}[t!]
	\centering
	\caption{The detailed statistics of three datasets}
	\label{tb-dataset}
	\begin{tabular}{ccrrrc}
		\hline
		\textbf{Dataset} & \textbf{Domain} & \textbf{Instances} & \textbf{Dimensions} & \textbf{Classes} & \textbf{Permissions} \\ \hline 
		PAMAP2 \cite{reiss2012introducing} & Activity Recognition & 1,140,000 & 27 & 18 & Public \\ 
		CAP \cite{terzano2002atlas} & Sleep Stage Detection & 921,700,000 & 21 & 8 & Public \\ 
		Mental Fatigue Dataset \cite{zhang2018deep} & Fatigue Detection & 1,458,648 & 4 & 2 &Private \\ \hline
	\end{tabular}
\end{table*}

\subsection{Inference}
Given samples as training dataset ${X=\{x_1,x_2,...,x_D\}}, x_i \in \mathbb{R}^{N \times T}$, we are able to compute the corresponding decision threshold ($\operatorname{THR}$):
\begin{equation}
\label{z}
\operatorname{THR} = \frac 1 D \sum_{i=1}^{D} \operatorname{Err} (x_i)+\sqrt{\frac 1 D \sum_{i=1}^{D}{(\operatorname{Err}(x_i)-\mu)^2}},
\end{equation}
where we denote $\mathrm{Err}(x_i)$ as the sum of loss function for $x_i$, and $\mu$ is the average value of $\mathrm{Err}(x_i)$ for $i=1,...,D$. The setting is similar to the normal training distribution $\mathcal N (\mu,\sigma)$ following with 1 standard deviation $\sigma$ of the mean $\mu$.

In inference process, the decision rule is that if $\mathrm{Err}(x_i)>\mathrm{THR} $, the testing sample in a sequence can be predicted to be ``abnormal'', otherwise ``normal''.

\renewcommand{\algorithmicrequire}{\textbf{Input:}}  
\renewcommand{\algorithmicensure}{\textbf{Output:}}  
\begin{algorithm}[h]
   \caption{Training and Inference procedure of CAE-M}
   \centerline{\textbf{Training process}} 
    \begin{algorithmic}[1]
    \REQUIRE  Normal Dataset $X=\{x_1,x_2,...,x_D\}$, time steps $h$, batch size $M$ and hyperparameters $\lambda_1, \lambda_2, \lambda_3$.
    \ENSURE Anomaly decision threshold THR and model parameter $w$.
    \STATE Transform each sample $x \in \mathbb{R}^{N \times T}$ into $x \in \mathbb{R}^{h \times N \times t}$ in the time axis;
    \STATE Randomly initialize parameter $w$;
    \WHILE{not converge} 
        % \hspace*{0.6cm} Sample batch $X_m$ from $X$;
        \STATE Calculate low-dimensional representation $z_f$ and reconstruction error $z_r$ at each time step; // Eq.~\eqref{z1} ~\eqref{eq-mse}
        \STATE Calculate MMD between $z_a$ and Gaussian distribution $P_z$; // Eq. (5)
        \STATE Combine $z_f$ and $z_r$ into $z_h = [z_f,z_r]_h$ for each sample; // Eq. (6)
        \STATE Predict the current value $z_h$ for the past values $[z_1,z_2,...,z_{h-1}]$ by Attention-based BiLSTM and AR model; // Eq. (7-18)
        \STATE Update $w$ by minimizing the compound objective function; // Eq. (20)
    \ENDWHILE
    \STATE Calculate the decision threshold THR by the training samples; // Eq.(21)
    \RETURN Optimal $w$ and THR.
    \end{algorithmic}

   \centerline{\textbf{Inference process}}
   \begin{algorithmic}[1]
   \REQUIRE  Normal and Anomalous dataset $X=\{x_1,x_2,...,$ $x_D\}$, threshold THR,  model parameter $w$, hyperparameters $\lambda_1$, $\lambda_2$ and $\lambda_3$.
   \ENSURE Label of all $x_i$.
    \FOR{\textbf{all} $x_i$ } 
    \STATE Calculate the loss $Err(x_i)=f(x_i;w)$; //$f(\cdot)$ denotes CAE-M
    \IF{$Err(x_i)>$ THR} 
    \STATE $x_i$ = ``anomaly'';
    \ELSE
    \STATE $x_i$ = ``normal'';
    \ENDIF
    \ENDFOR
    \RETURN Label of all $x_i$.
    \end{algorithmic} 
    \label{algorithm1}
\end{algorithm}

The complete training and inference procedure of CAE-M is shown in Algorithm \ref{algorithm1}.

\section{Experiments}
\label{sec:exp}

In this section, we conduct extensive experiments to evaluate the performance of our proposed CAE-M approach for anomaly detection on several real-world datasets.

\subsection{Datasets}
%正常有很多种可能，异常也有很多种可能，比如检测人是不是疲劳，正常人不疲劳但表现也不同，疲劳的人程度也不同，但是唯一能确定的是正常类是占多数类，异常类别占少数类

We adopt two large publicly-available datasets and a private dataset: PAMAP2, CAP and Mental fatigue dataset. These datasets are exploiting multi-sensor time series for activity recognition, sleep state detection, and mental fatigue detection, respectively. Therefore, they are ideal testbeds for evaluating anomaly detection algorithms.

\textbf{PAMAP2}~\cite{reiss2012introducing} dataset is a mobile dataset with respect to actions or activities from UCI repository, containing data of 18 different physical activities performed by 9 subjects wearing 3 inertial measurement units, e.g. accelerator, gyroscope and magnetometer. There are 18 activity categories in total. For experiments, we treat these classes with relatively smaller samples as the anomaly classes (including running, ascending stairs, descending stairs and rope jumping), while the rest categories are combined to form the normal classes.

\textbf{CAP Sleep Database}~\cite{terzano2002atlas}, which stands for the Cyclic Alternating Pattern (CAP) database, is a clinical dataset from PhysioNet repository. It is characterized by periodic physiological signals occurring during wake, S1-S4 sleep stages and REM sleep. The waveforms include at least 3 EEG channels, 2 EOG channels, EMG signal, respiration signal and EKG signal. There are 16 healthy subjects and 92 patients in the database. The pathological recordings include the patients diagnosed with bruxism, insomnia, narcolepsy, nocturnal frontal lobe epilepsy, periodic leg movements, REM behavior disorder and sleep-disordered breathing. In this task, we extracted 7 valid channels of all the channels like ROC-LOC, C4-P4, C4-A1, F4-C4, P4-O2, ECG1-ECG2, EMG1-EMG2 etc. For detecting sleep apnea events, we chose healthy subjects as normal class and the patients with sleep-disordered breathing as anomaly class.

\textbf{Mental Fatigue Dataset}~\cite{zhang2018deep} is a real world health-care dataset. Aiming to detect mental fatigue in the healthy group, we collected the physiological signals (e.g., GSR, HR, R-R intervals and skin temperature) using wearable device. There are 6 healthy young subjects participated in the mental fatigue experiments. In this task, non-fatigue data samples are labeled as normal class and fatigue data samples are labeled as anomaly class. Fatigue data accounts for a fifth of the total. 

The detailed information of the datasets is shown in \tablename~\ref{tb-dataset}.

\subsection{Baseline Methods}
In order to extensively evaluate the performance of the proposed CAE-M approach, we compare it with several traditional and deep anomaly detection methods:

(1)~\textbf{KPCA} (Kernel principal component analysis)~\cite{hoffmann2007kernel}, which is a non-linear extension of PCA commonly used for anomaly detection. (2)~\textbf{ABOD} (Angle-based outlier detection)~\cite{kriegel2008angle}, which is a probabilistic model that well suited for high dimensional data. (3)~\textbf{OCSVM} (One-class support vector machine)~\cite{ma2003time}, which is the one-class learning method that classifies new data as similar or different to the training set. (4)~\textbf{HMM} (Hidden Markov Model)~\cite{joshi2005investigating} is a finite set of states, each of which is associated with a probability distribution. In a particular state an observation can be generated, according to the associated probability distribution. (5)~\textbf{CNN-LSTM}~\cite{donahue2015long}, which is a forecasting model composed of convolutional and LSTM networks. It can obtain the forecast by estimating the current data, and detect anomalies on comparing the forecasting value with actuals. (6)~\textbf{LSTM-AE} (LSTM based autoencoder)~\cite{malhotra2016lstm}, which is an unsupervised detection technique used in time series that can induce a representation by learning an approximation of the identity function of data. (7)~\textbf{ConvLSTM-COMPOSITE}~\cite{medel2016anomaly}, which utilizes a composite structure that is able to encoder the input, reconstruct it, and predict its near future. To simplify 
the name, ``ConvLSTM-COMP'' denotes ConvLSTM-COMPOSITE. We choose the ``conditional'' version to build a single model called \textbf{ConvLSTM-AE} by removing the forecasting decoder. (8)~\textbf{UODA} (Unsupervised sequential outlier detection with deep architecture)~\cite{lu2017unsupervised}, which utilizes autoencoders to capture the intrinsic difference between normal and abnormal samples, and then integrates the model to RNNs that perform fine-tuning to update the parameters in DAE. (9)~\textbf{MSCRED} (Multi-scale convolutional recurrent encoder-decoder)~\cite{zhang2019deep}, which is a reconstruction-based anomaly detection and diagnosis method. 

\renewcommand{\arraystretch}{1.5} %控制行高  
\begin{table*}[!t]
	
	\centering  
	\fontsize{7.2}{8}\selectfont  
	\begin{threeparttable}  
		\caption{The mean precision, recall and F1 score of baselines and our proposed method, * p-value  = 0.0077.}  
		\label{tab:performance_comparison}  
		\setlength{\tabcolsep}{4.5mm}{
		\begin{tabular}{cccccccccc}  
			\toprule  
			\multirow{2}{*}{Method}&  
			\multicolumn{3}{c}{PAMAP2}&\multicolumn{3}{c}{CAP dataset}&\multicolumn{3}{c}{Fatigue dataset}\cr  
			\cmidrule(lr){2-4} \cmidrule(lr){5-7}\cmidrule(lr){8-10}    
			&mPre&mRec&mF1&mPre&mRec&mF1&mPre&mRec&mF1\cr  
			\midrule  
			KPCA&0.7236&0.6579&0.6892&0.7603&0.5847&0.6611&0.5341&0.5014&0.5173\cr  
			ABOD&0.8653&0.9022&0.8834&0.7867&0.6365&0.7037&0.6679&0.6145&0.6401\cr  
			OCSVM&0.7600&0.7204&0.7397&0.9267&0.9259&0.9263&0.5605&0.5710&0.5290\cr  
			HMM&0.6950&0.6553&0.6745&0.8238&0.8078&0.8157&0.6066&0.6076&0.6071\cr  
			\hline  
			  
			CNN-LSTM&0.6680&0.5392&0.5968&0.6159&0.5217&0.5649&0.5780&0.5042&0.5386\cr  
			LSTM-AE&0.8619&0.7997&0.8296&0.7147&0.6253&0.6671&0.7140&0.6820&0.6870\cr 
			UODA&0.8957&0.8513&0.8730&0.7557&0.5124&0.6107&0.8280&0.7770&0.8017\cr  
			MSCRED&0.6997&0.7301&0.7146&0.6410&0.5784&0.6081&0.8016&0.6802&0.7359\cr 
			ConvLSTM-AE&0.7359&0.7361&0.7360&0.8150&0.8194&0.8172&0.9010&0.9346&0.9175\cr  
			ConvLSTM-COMP&0.8844&0.8842&0.8843&0.8367&0.8377&0.8372&0.9373&0.9316&0.9344\cr 
			  
			\hline 
			CAE-M (Ours)&{\bf 0.9608}&{\bf 0.9670}&{\bf 0.9639}&{\bf 0.9939}&{\bf 0.9952}&{\bf 0.9961}&{\bf 0.9962}&{\bf 0.9959}&{\bf 0.9960}\cr 
			Improvement &{ 7.64\%}&{ 6.48\%}&{7.96\%}&{ 6.72\%}&{6.93\%}&{6.98\%}&{5.89\%}&{6.13\%}&{ 6.16\%}\cr
			\bottomrule  
		\end{tabular}  }
	\end{threeparttable}  
\end{table*}  

\subsection{Implementation details}

For traditional anomaly detection, we scale the sequential data into segments and extract the features from each segment. In PAMAP2 dataset, multiple sensors are worn on three different position (wrist, chest, ankle). Hence, we extract 324 features including time and frequency domain features. In CAP Sleep dataset, we first pass through the Hanning window low pass filter for removing the high frequency components of signals. And then we extract 91 features for EEG, EMG and ECG signals \cite{subha2010eeg,phinyomark2009novel,gautama2017overview}; In Mental Fatigue dataset, we preprocess physiological signals by interpolation and filtering algorithm. Then we extract 23 features for Galvanic Skin Response (GSR), Heart Rate (HR), R-R intervals and skin temperature sensors \cite{zhang2018deep}.

For Deep Anomaly Detection (DAD) method, we filter multi-sensor signals and then pack these signals into matrix as input to construct the deep model.

We reimplement these methods based on open-source repositories\footnote{\url{https://pyod.readthedocs.io/en/stable/}, \url{https://github.com/7fantasysz/MSCRED}} or our own implementations. For KPCA, we employ Gaussian kernel with a bandwidth of 600, 500, 0.5, respectively for PAMAP2, CAP, and Mental Fatigue datasets. For ABOD, we use $k$ nearest neighbors to approximate the complexity reduction. For an observation, the variance of its weighted cosine scores to all neighbors could be viewed as the abnormal score. For OCSVM, we adopt PCA for OCSVM as a dimension reduction tool and employ the Gaussian kernel with a bandwidth of 0.1. For HMM, we build a Markov model after extracting features and calculate the anomaly probability from the state sequence generated by the model. For CNN-LSTM, we define a CNN-LSTM model in \texttt{Keras} by first defining 2D convolutional network as comprised of Conv2D and MaxPooling2D layers ordered into a stack of the required depth, wrapping them in a TimeDistributed layer and then defining the LSTM and output layers. For LSTM-AE, we use single-layer LSTM on both encoder and decoder in the task. For ConvLSTM-COMPOSITE, we choose "conditional" version and adapt this technique to anomaly detection in multivariate time series. Here we also build a single model called ConvLSTM-AE by removing forecasting decoder. For UODA, we reimplement this algorithm by customizing the number of layers and hyper-parameters. For MSCRED, we first construct multi-scale matrices for multi-sensor data, and then fed it into MSCRED model and evaluate the performance. 

For our own CAE-M, we use library Hyperopt \cite{bergstra2013hyperopt} to select the best hyper-parameters (i.e., time window, the number of neurons, learning rate, activation function, optimization criteria and iterations). The characterization network runs with $Conv2D$ $\rightarrow$ $ Maxpooling$ $\rightarrow$ $Conv2D$ $\rightarrow$ $Maxpooling$ $\rightarrow$ $Conv2DTranspose$ $\rightarrow$ $Conv2DTranspose$ $\rightarrow$ $Conv2DTranspose$, i.e., Conv1-Conv5 with 32 kernels of size 4 $\times$ 4, 64 kernels of size 4 $\times$ 4, 64 kernels of size 4 $\times$ 4, 32 kernels of size 4 $\times$ 4, 1 kernels of size 4 $\times$ 4, and Maxpooling with size 2 $\times$ 2. We use Rectified Linear Unit (ReLU) as the activation function of convolutional layers. The memory network contains non-linear prediction and linear prediction, where the non-linear network runs with $BiLSTM(512)$ $\rightarrow$ $Attention(h-1)$ $\rightarrow$ $Dropout(0.2)$ $\rightarrow$ $FC(1000,linear)$, and the linear network runs with $FC(1000,linear)$. The CAE-M model is trained in an end-to-end fashion using \texttt{Keras} \cite{ketkar2017introduction}. The optimization algorithm is Adam and the batch size is set as 32. And we set parameters of compound objective function $\lambda_1=e-04$, $\lambda_2=0.5$ and $\lambda_3=0.5$. The time step $h$ usually gives desirable results as $h=5$ or $h=10$.

Note that in addition to the complete CAE-M approach, we further evaluate its several variants as baselines to justify the effectiveness of each component:
\begin{itemize}
	\item CAE-M\textit{w/o}Pre. The CAE-M model removes the linear and non-linear prediction. That is, this variant only adopts the characterization network with reconstruction loss and MMD loss. (i.e., $\lambda_1=e-04, \lambda_2=0, \lambda_3=0$)
\end{itemize}
\begin{itemize}
	\item CAE-M\textit{w/o}Rec+MMD. The CAE-M model removes the reconstruction error and MMD. Different from CNN-LSTM model, the characterization network is still performed as the deep convolutional autoencoder. We put the latent representation without reconstruction error into the memory network. (i.e., $\lambda_1=0, \lambda_2=0.5, \lambda_3=0.5$)
\end{itemize}
\begin{itemize}
	\item CAE-M\textit{w/o}ATTENTION. The CAE-M model without Attention component is implemented. (i.e., $\lambda_1=e-04, \lambda_2=0.5, \lambda_3=0.5$)
	%除了MMD,都加的基础上，有和没有双向lstm+attention
\end{itemize}
\begin{itemize}
	\item CAE-M\textit{w/o}AR. The CAE-M model without AR component is implemented. (i.e., $\lambda_1=e-04, \lambda_2=0.5, \lambda_3=0$)
	%除了MMD，都加的基础上，有和没有AR
\end{itemize}
\begin{itemize}
	\item CAE-M\textit{w/o}MMD. The CAE-M model without MMD component is implemented. (i.e., $\lambda_1=0, \lambda_2=0.5, \lambda_3=0.5$)
	%都加的基础上，有和没有MMD
\end{itemize}

Note that anomaly detection problems are often with highly-imbalanced classes, hence \textit{accuracy} is not suitable as the evaluation metric. In order to thoroughly evaluate the performance of our proposed method, we follow existing works \cite{zong2018deep,guo2018multidimensional,zhang2019deep} to adopt the mean \textit{precision, recall}, and \textit{F1} score as the evaluation metrics. The mean \textit{precision} means the average 
precision of normal and abnormal class. The same pattern goes for mean \textit{recall, F1} score.

In the experiments, the train-validation-test sets are split by following existing works \cite{zhang2019deep,lu2017unsupervised}. Concretely speaking, for each dataset, we split normal samples into training, validation, and test with the ratio of $5:1:4$, where the training and validation set only contain normal samples and have no overlapping with testing set. The anomalous samples are only used in the testing set. The model selection criterion, i.e., hyperparameters, used for tuning is the validation error on the validation set.

\renewcommand{\arraystretch}{1.5}
\begin{table*}[!t]  
	
	\centering  
	\fontsize{7.2}{8}\selectfont  
	\caption{Detection performance in different sleep stages of baselines and our proposed method, *p-value = 0.0074.}  
	\label{tab:performance}  
	\setlength{\tabcolsep}{1.1mm}{{
		\begin{tabular}{ccccccccccccccccccc}  
			\toprule  
			\multirow{2}{*}{Method}&  
			\multicolumn{3}{c}{WAKE}&\multicolumn{3}{c}{S1}&\multicolumn{3}{c}{S2}&\multicolumn{3}{c}{S3}&\multicolumn{3}{c}{S4}&\multicolumn{3}{c}{REM}\cr \cline{2-19}  
			&mPre&mRec&mF1&mPre&mRec&mF1&mPre&mRec&mF1&mPre&mRec&mF1&mPre&mRec&mF1&mPre&mRec&mF1\cr  
			\hline  
			  
			KPCA&0.9162&0.8213&0.8662&0.8267&0.7598&0.7918&0.9257&0.9353&0.9305&0.9039&0.8689&0.8861&0.9402&0.9604&0.9502&0.9536&0.9614&0.9575\cr  
			ABOD&0.9872&0.8686&0.9242&0.9347&0.5522&0.6942&0.9389&0.6550&0.7716&0.8489&0.6184&0.7155&0.6749&0.6448&0.6595&0.5915&0.5909&0.5912\cr  
			OCSVM&0.9784&0.9492&0.9636&0.9655&0.9504&0.9579&0.9395&0.9448&0.9421&{\bf0.9714}&{\bf0.9499}&{\bf0.9605}&0.8701&0.9488&0.9077&{\bf0.9784}&0.9492&0.9636\cr  
			HMM&0.8417&0.8406&0.8411&0.8790&0.8856&0.8823&0.8967&0.8887&0.8927&0.6880&0.6747&0.6813&0.7279&0.7286&0.7282&0.8024&0.8649&0.8325\cr\hline  
			LSTM-AE&0.6990&0.7178&0.7082&0.6517&0.6492&0.6504&0.7430&0.7331&0.7380&0.7689&0.7828&0.7758&0.7274&0.7569&0.7418&0.6590&0.6887&0.6735\cr 
			UODA&0.6159&0.6326&0.6241&0.6762&0.6762&0.6762&0.7290&0.5223&0.6086&0.5716&0.5766&0.5741&0.6626&0.8498&0.6807&0.5626&0.6116&0.5861\cr  
			\makecell*[r]{ConvLSTM- \\ \scriptsize COMP} &0.9889&0.9772&0.9830&0.9755&0.9850&0.9864&0.9250&0.9127&0.9188&0.9401&0.9041&0.9217&0.8647&0.8866&0.9023&0.9675&0.9949&0.9810\cr\hline 
			CAE-M&{\bf 0.9974}&{\bf 0.9949}&{\bf 0.9961}&{\bf 0.9958}&{\bf 0.9950}&ß{\bf 0.9954}&{\bf 0.9950}&{\bf 0.9950}&{\bf 0.9950}&0.9294&0.8842&0.9063&{\bf 0.9842}&{\bf 0.9950}&{\bf0.9895}&0.9681&{\bf 0.9950}&{\bf 0.9813}\cr  
			\bottomrule  
		\end{tabular}  
	}}
\end{table*}

\subsection{Results and Analysis}
%et the precision and recall for each class and average, as you said. 
%http://nooverfit.com/wp/david9%E7%9A%84%E6%99%AE%E5%8F%8A%E8%B4%B4%EF%BC%9A%E6%9C%BA%E5%99%A8%E8%A7%86%E8%A7%89%E4%B8%AD%E7%9A%84%E5%B9%B3%E5%9D%87%E7%B2%BE%E5%BA%A6ap-%E5%B9%B3%E5%9D%87%E7%B2%BE%E5%BA%A6%E5%9D%87/
As shown in \tablename~\ref{tab:performance_comparison}, we compare our proposed method with traditional and deep anomaly detection methods using the mean precision, recall and F1 score. We can see that our method outperforms most of the existing methods, which demonstrates the effectiveness of our method. From \tablename~\ref{tab:performance_comparison}, we can observe the following results.

For the PAMAP2 dataset, the CAE-M achieves the highest precision and recall compared by 10 popular methods. Traditional methods perform differently on PAMAP2 dataset since they are limited by the feature extraction and feature selection methods. In deep learning method, CNN-LSTM has a lowest F1 score. This means that more constraints such as data preprocessing method and anomaly evaluation strategy need to be added for prediction-based anomaly detection. For LSTM-AE, MSCRED and ConvLSTM-AE, they both are reconstruction-based anomaly detection methods. Their performance is limited by the ``noisy data'' problem, resulting in reconstruction error for the abnormal input could be fit so well. For UODA, it performs reasonably well on the PAMAP2 dataset, but it is not end-to-end training, which is needed by pre-training denoising autoencoder (DAEs) and deep recurrent networks (RNNs), and then fine-tuning the UODA model composing of the DAE and RNN. For ConvLSTM-COMPOSITE model, it performs better than other baseline models. The model consists of a single encoder, two decoders of reconstruction branch and prediction branch. In fact, since its efficiency is influenced by reconstruction error and prediction error respectively, its performance could be limited by one of encoder-decoder models.

For the CAP dataset, most of methods show a low F1 score. As CAP dataset contains different sleep stages of subjects, some methods are limited by high complexity of data. For OCSVM and HMM, they achieve better performance because of dimensionality reduction from 36 dimensions of PAMAP2 dataset to 7 dimensions.  For MSCRED, due to \textit{batch size =1} for the training model in the open source code, the loss function couldn't converge during training model and the training speed is slow. Our proposed method achieves about 7\% improvement at F1 score, compared with the existing methods.

For Fatigue dataset, it is difficult to label fatigue and non-fatigue data manually. Therefore, it may be a lot of noise or misclassification patterns in the data, so that most of methods fail to solve this problem. For UODA, MSCRED and ConvLSTM, they have ability to overcome noise and misclassification of training data. Our proposed method also solves this problem successfully and achieves at least 6\% improvement at F1 score.

Besides, in order to indicate significant differences from our proposed method and other baselines, we use Wilcoxon signed rank test\cite{0Wilcoxon} to analyze these results in \tablename~\ref{tab:performance_comparison}. We compute average p-value of CAE-M compared with other baselines. A p-value = 0.0077 indicates that the performance of our proposed method differs from other methods. This p-value is also computed in \tablename~\ref{tab:performance}.
%In sum, the CAE-M model achieves the best accuracy on the public or private datasets.

\subsection{Fine-grained Analysis}

In addition to the anomaly detection of different classes on each dataset, we conduct a fine-grained analysis to evaluate the performance of each method within each class. Considering intra-class diversity, we conduct a group of experiments to detect anomalies in different sleep stages. In fact, these physiological signals in different sleep stages have significant differences. We choose 4 traditional methods and 3 deep methods with good performance in global domain as comparison methods. As shown in \tablename~\ref{tab:performance}, we can observe that our architecture is most robust across same experiment settings. Several observations from these results are worth highlighting. For ABOD, the testing performance is unstable in local domain, which the highest F1 score is 0.92 in WAKE and the lowest F1 score is 0.59 in REM. For KPCA and ConvLSTM-COMPOSITE, the testing performance in local domain far exceeds the performance in global domain. This demonstrates that the two model can achieve better performance when intra-class data have similar distribution or regular pattern. For other methods, the testing performance is consistent in local and global domain. For our proposed method, the best testing performance can be achieved no matter in local domain or global domain. This study clearly justifies the superior representational capacity of our architecture to solve intra-class diversity.

\subsection{Effectiveness Evaluation}

\begin{table}[t!]
	\centering
	\caption{The evaluation results on LOSO cross validation approach, including the best, the worst and the mean F1 score of 8 subjects.}  
	\label{tab:LOSO1} 
	\fontsize{8}{8}\selectfont  
	\begin{tabular}{cccc}
		\toprule
		Method & Worst mF1 & Best mF1 & Mean mF1 \\ \hline 
		ABOD               & 0.6093  & 0.8507   & 0.7706   \\ 
		ConvLSTM-COMP & 0.7033  & 0.9224   & 0.8493   \\ 
		UODA               & 0.5938  & 0.9336   & 0.7984   \\ 
		CAE-M              & \textbf{0.8009}  & \textbf{0.9433}   & \textbf{0.8616}   \\ \bottomrule
	\end{tabular}
\end{table} 

\subsubsection{Leave One Subject Out}
\begin{figure*}
	\centering 
	\subfigure[Leave One Subject Out Evaluation]{\includegraphics[width=3in]{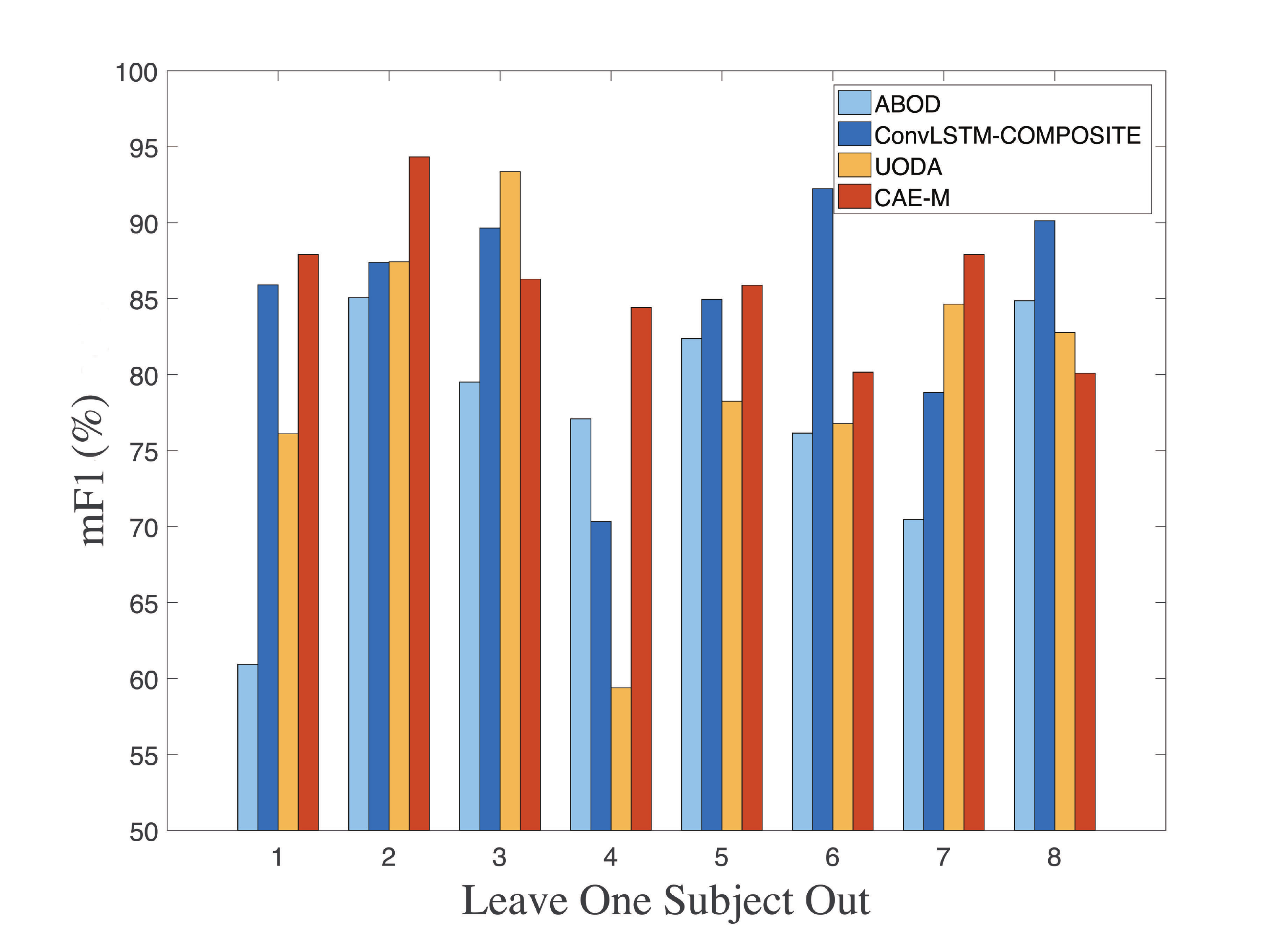}
	\label{fig-loso}}
	\subfigure[Ablation Study]{\includegraphics[width=3in]{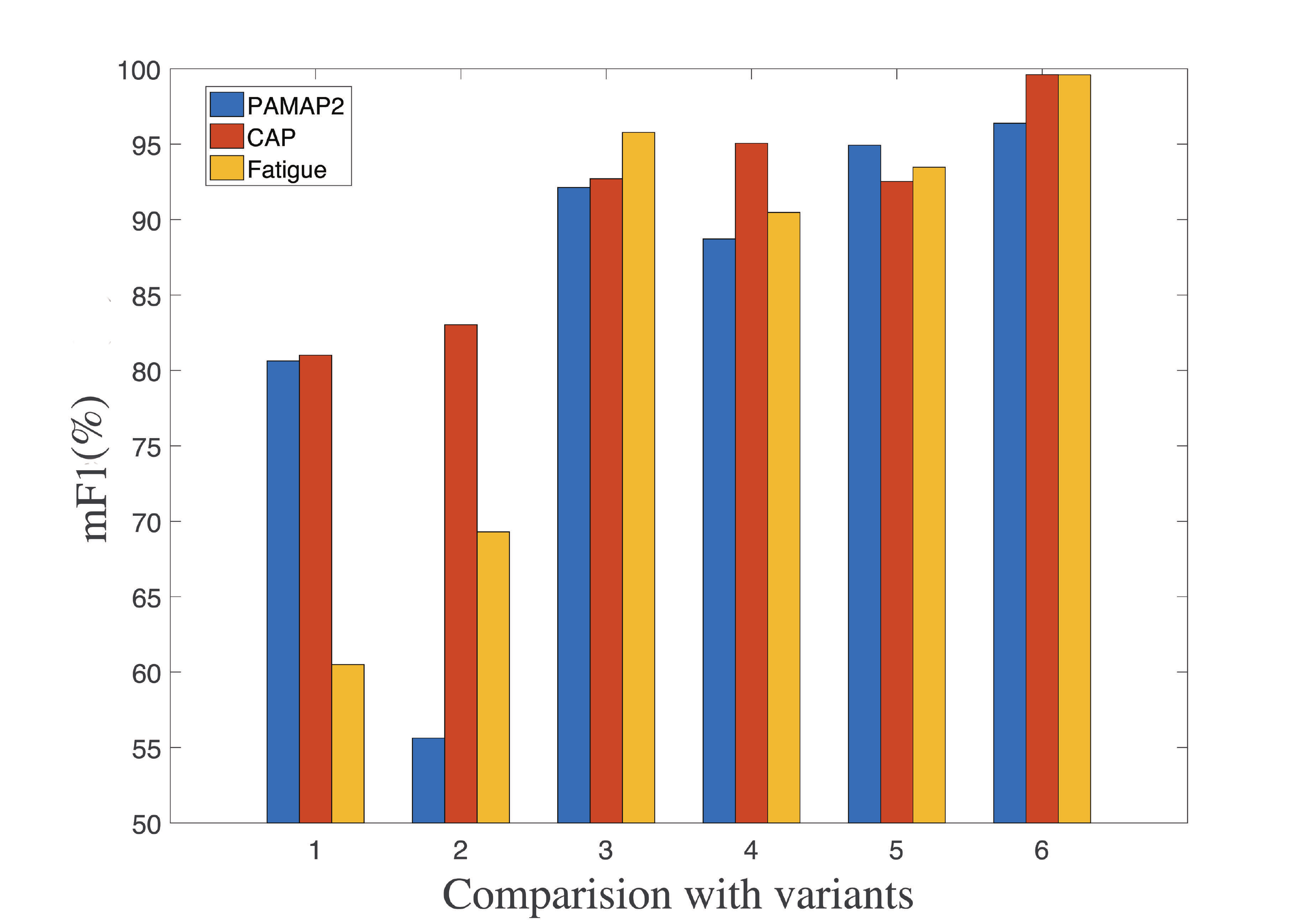}
	\label{fig-ablation2}}
	\caption{Effectiveness evaluation using LOSO method and ablation study.}
	\label{fig-ablation}
\end{figure*}

In this section, we measure the generalization ability of models using Leave One Subject Out (LOSO). The fact is that when training and testing datasets contain the same subject, the model is likely to know more about the current subject which may be biased towards a new one. Therefore, LOSO could help to evaluate the generalization ability. We choose the PAMAP2 dataset to conduct subject-independent experiments which contain 8 subjects. As can be seen in \figurename~\ref{fig-loso}, we evaluate our proposed method and three methods with relatively high F1 score. By examining the results, one can easily notice that deep learning-based methods obtain better performance than traditional methods. However, complex models such as deep neural networks are prone to overfitting because of their flexibility in memorizing the idiosyncratic patterns in the training set, instead of generalizing to unseen data. 

\tablename~\ref{tab:LOSO1} shows the best, the worst and average performance among 8 subjects. We can observe that UODA and ConvLSTM-COMPOSITE model perform well in some specific subjects, but they fail to reduce the effects of overfitting to each test subject, even drop to 0.70 and 0.59 for some subjects. Compared to these methods, CAE-M can generalize well on testing subjects it hasn't appeared before, which reach the average F1 score of 0.86. Besides, we perform an analysis of variance on repeated measures within subject 1 (corresponding to numbers in \figurename~\ref{fig-loso}). As shown in \tablename~\ref{tab:LOSO2}, we observe that CAE-M remains a more stable performance on repeated measurements. In summary, the above demonstrates that our model can be motivated to improve the generalization ability.

\begin{table}[t!]
	\centering
	\caption{The repeated measures analysis of variance on LOSO cross validation approach of one subject.}  
	\label{tab:LOSO2} 
	\fontsize{7.7}{8}\selectfont 
	\begin{tabular}{cccc}
		\toprule
		Method &mPre &mRec & mF1 \\ \hline 
		ABOD               &  0.6240$\pm$0.000  & 0.5946$\pm$0.000   &0.6090$\pm$0.000   \\ 
		ConvLSTM-COMP & 0.8953$\pm$0.029  & 0.8081$\pm$0.038 & 0.8488$\pm$0.019   \\ 
		UODA               & 0.8155$\pm$0.063 & 0.7464$\pm$0.027 & 0.7782$\pm$0.031   \\ 
		CAE-M              & \textbf{0.9437$\pm$0.024}  & \textbf{0.8191$\pm$0.003} & \textbf{0.8770$\pm$0.012}   \\ \bottomrule
	\end{tabular}
\end{table}

%Leave one out subject makes it sure that you don't have subject bias.

%The fact that you have the same subject in your training and your testing datasets will make the model know more about your subject than it should. With a brand new subject, the model will probably perform poorly because it never trained on the subject before.

\subsubsection{Ablation Study}
\renewcommand{\arraystretch}{1.5} %控制行高  
\begin{table*}[!t]
	
	\centering  
	\fontsize{7.2}{8}\selectfont  
	\begin{threeparttable}  
		\caption{The mean precision, recall and F1 score from variants.}  
		\label{tab:variant}  
		\setlength{\tabcolsep}{4mm}{
		\begin{tabular}{ccccccccccc}  
			\toprule  
			\multirow{2}{*}{ID}&
			\multirow{2}{*}{Method}&  
			\multicolumn{3}{c}{PAMAP2}&\multicolumn{3}{c}{CAP dataset}&\multicolumn{3}{c}{Fatigue dataset}\cr  
			\cmidrule(lr){3-5} \cmidrule(lr){6-8}\cmidrule(lr){9-11}    
			&&mPre&mRec&mF1&mPre&mRec&mF1&mPre&mRec&mF1\cr  
			\midrule  
			1&CAE-M$_ \textit{w/o Pre} $&0.8103&0.8023&0.8063&0.8299&0.8101&0.8199&0.6005&0.6096&0.6050\cr  
			2&CAE-M$_ \textit{w/o Rec+MMD} $&0.5693&0.5440&0.5563&0.8896&0.7784&0.8303&0.7050&0.6814&0.6930\cr  
			3&CAE-M$_ \textit{w/o ATTENTION} $&0.9151&0.9276&0.9213&0.9251&0.9291&0.9271&0.9605&0.9551&0.9578\cr 
			4&CAE-M$_ \textit{w/o AR} $&0.9060&0.8691&0.8872&0.9634&0.9381&0.9506&0.9046&0.9048&0.9047\cr  
			5&CAE-M$_ \textit{w/o MMD} $&0.9437&0.9550&0.9493&0.9293&0.9213&0.9253&0.9407&0.9288&0.9347\cr 
			\hline  
			\hline 
			6&CAE-M&{\bf 0.9608}&{\bf 0.9670}&{\bf 0.9639}&{\bf 0.9939}&{\bf 0.9952}&{\bf 0.9961}&{\bf 0.9962}&{\bf 0.9959}&{\bf 0.9960}\cr 
			\bottomrule  
		\end{tabular}  }
	\end{threeparttable}  
\end{table*} 
The proposed CAE-M approach consists of several components such as CAE, MMD, Attention mechanism, BiLSTM and Auto-regressive. To demonstrate the effectiveness of each component, we conduct ablation studies in this section. The ablation study is shown in \figurename~\ref{fig-ablation2}. These ID numbers represent CAE-M without non-linear and linear prediction, CAE-M without reconstruction error and MMD, CAE-M without attention module, CAE-M without AR, CAE-M without MMD and CAE-M, respectively. The experimental results indicate that for the removal of different component above, there is corresponding performance drop at F1 score. We can observe that CAE-M model without prediction or reconstruction error achieves a low F1 score relatively. This demonstrates that our composite model is effective and necessary for anomaly detection in multi-sensor time-series data. Compared to original CAE-M model, removing the AR component (in CAE-M$_ \textit{w/o AR}$) from the full model causes significant performance drops on most of the datasets. This shows the critical role of the AR component in general. Moreover, attention and MMD components can also cause big performance rises on all the datasets. More details are shown in \tablename~\ref{tab:variant}. Here, these ID numbers are corresponding to numbers in \figurename~\ref{fig-ablation2}.

\subsection{Robustness to Noisy Data}

\begin{figure}[t!]
	\centering 
	\subfigure[mF1]{\includegraphics[width=1.1in]{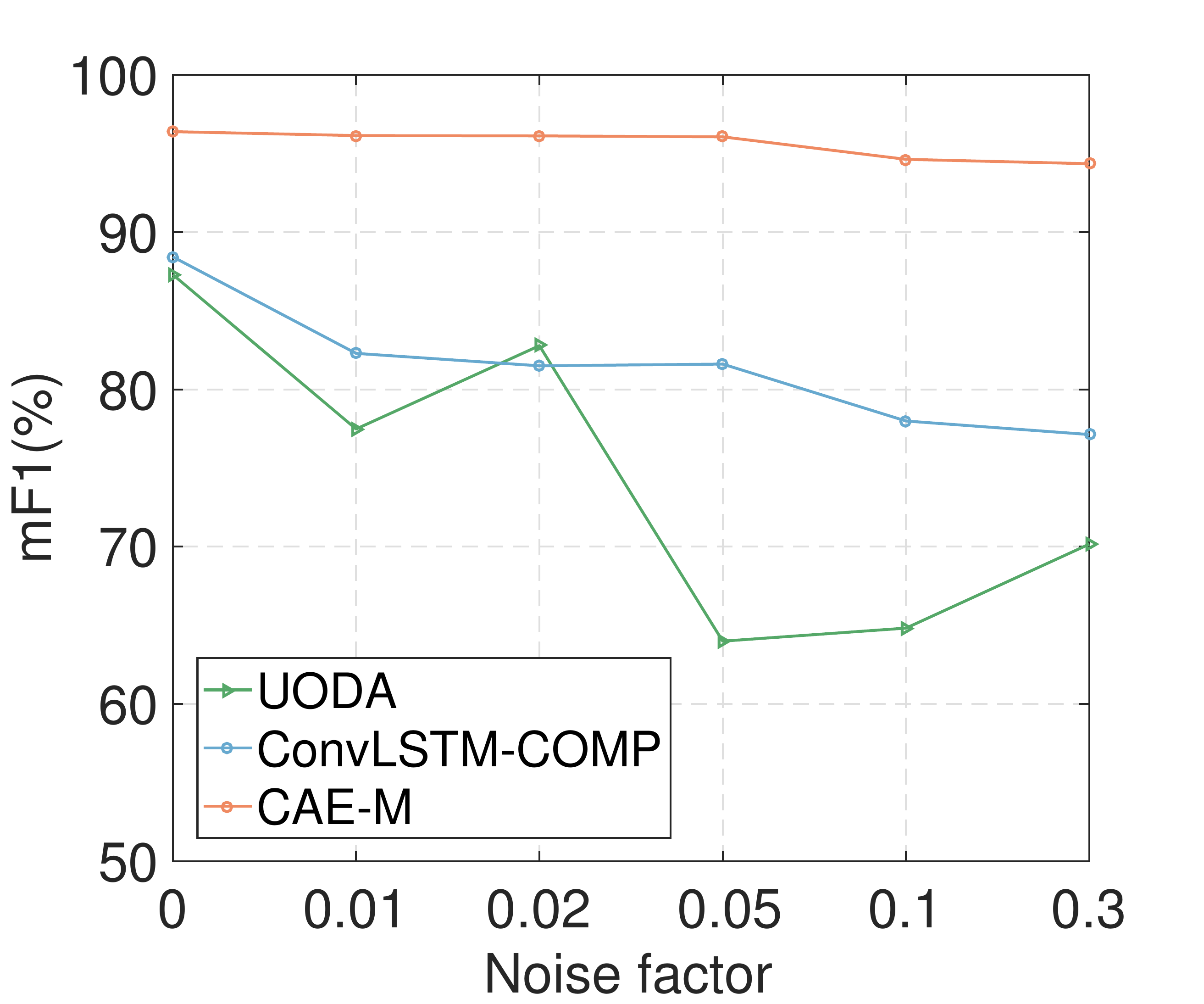}}
	\subfigure[mPre]{\includegraphics[width=1.1in]{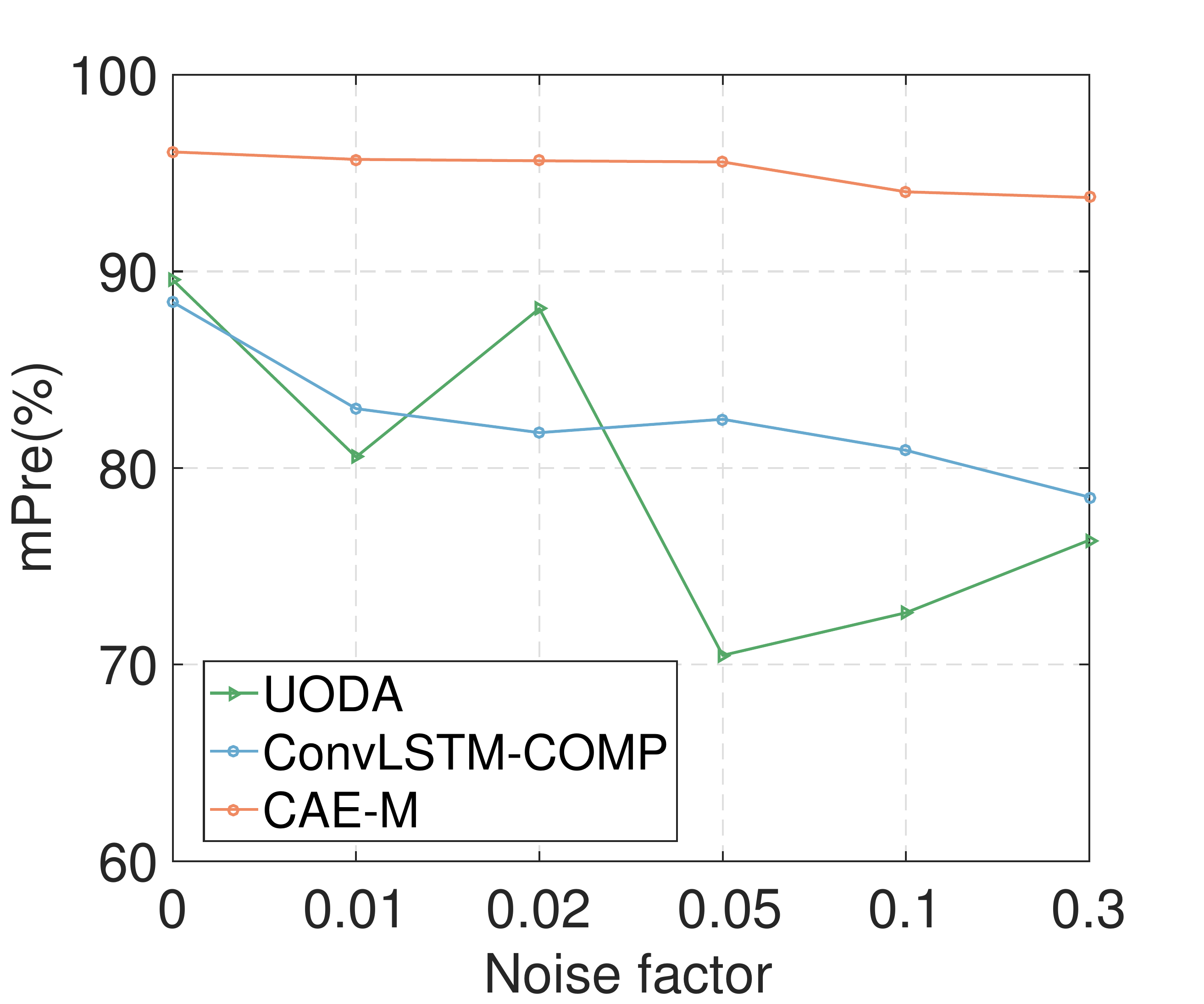}}
	\subfigure[mRec]{\includegraphics[width=1.1in]{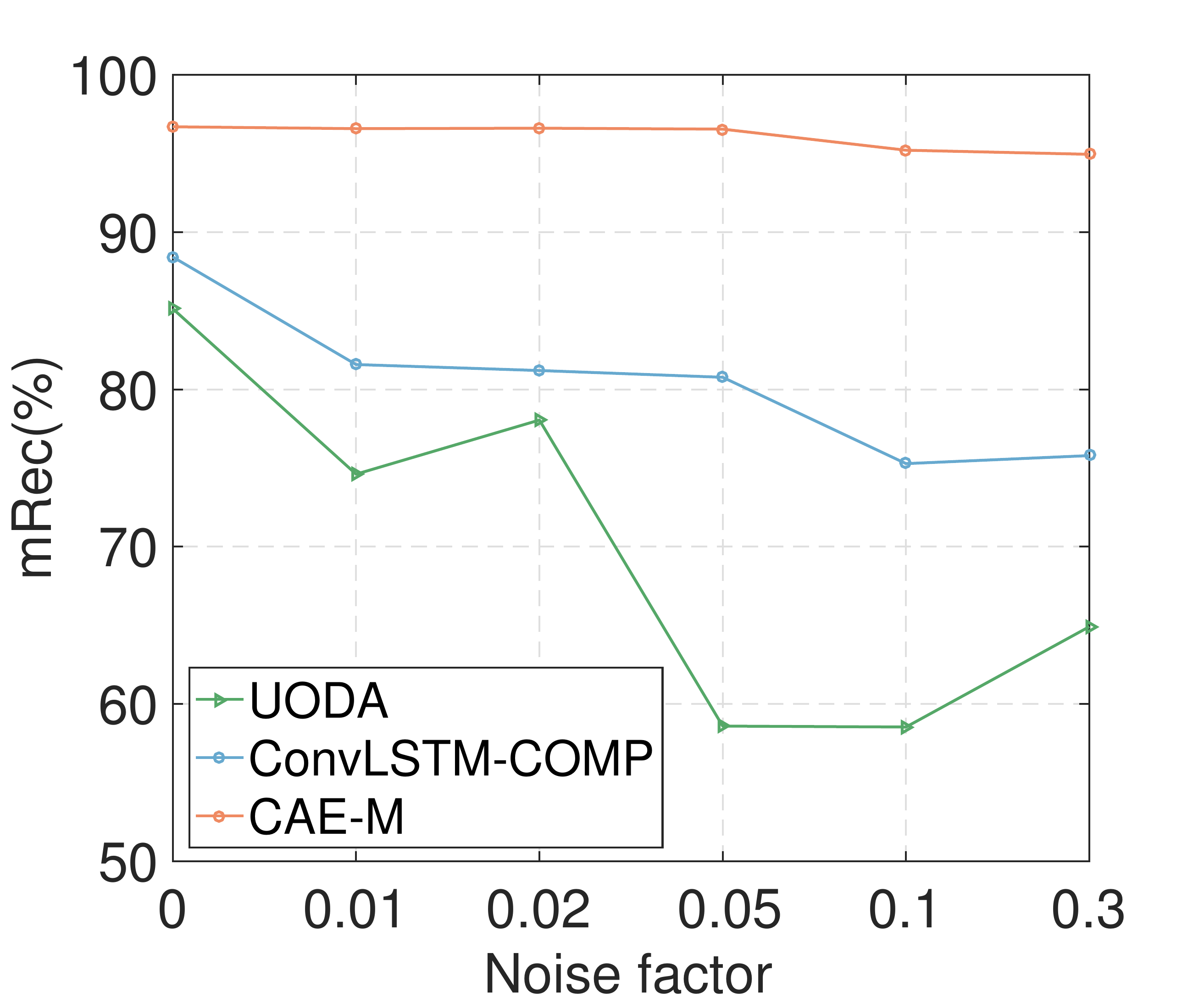}}
	\caption{Robustness to noisy data.}
	\label{fig:noise}
\end{figure}

In real-world applications, the collection of multi-sensor time-series data can be easily polluted with noise due to changes in the environment or the data collection devices.
The noisy data bring critical challenges to the unsupervised anomaly detection methods. In this section, we evaluate the robustness of different methods to noisy data.
We manually control the noisy data ratio in the training data. We inject Gaussian noise ($\mu$=0, $\sigma$=0.3) in a random selection of samples with a ratio varying between 1\% to 30\%. We compare the performance of three methods on PAMAP2 dataset: UODA, ConvLSTM-COMPOSITE, and CAE-M in \figurename~\ref{fig:noise}. These methods have good stability in the above experiments. As the noise increases, the performance of all methods decreases. For CAE-M, the F1 score, precision and recall have no significant decline. Among them, our model remains significantly superior to others, demonstrating its robustness to noisy data. 

\subsection{Further Analysis}

\subsubsection{Parameter Sensitivity Analysis}
\label{sec:parameter}

\begin{figure}[t!]
	\centering 
	\subfigure[MMD loss]{\includegraphics[width=1.12in]{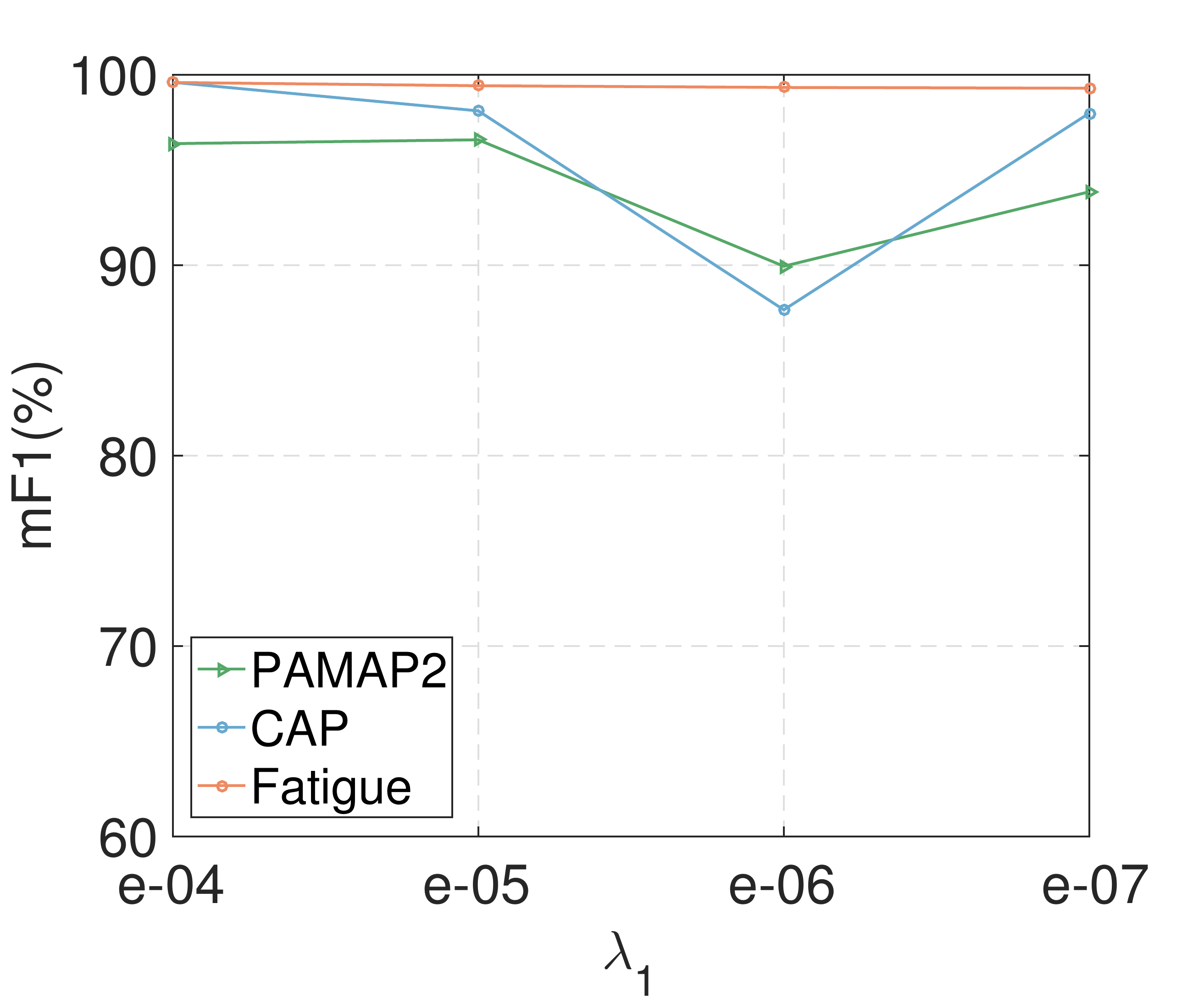}}
	\subfigure[Bi-LSTM loss]{\includegraphics[width=1.12in]{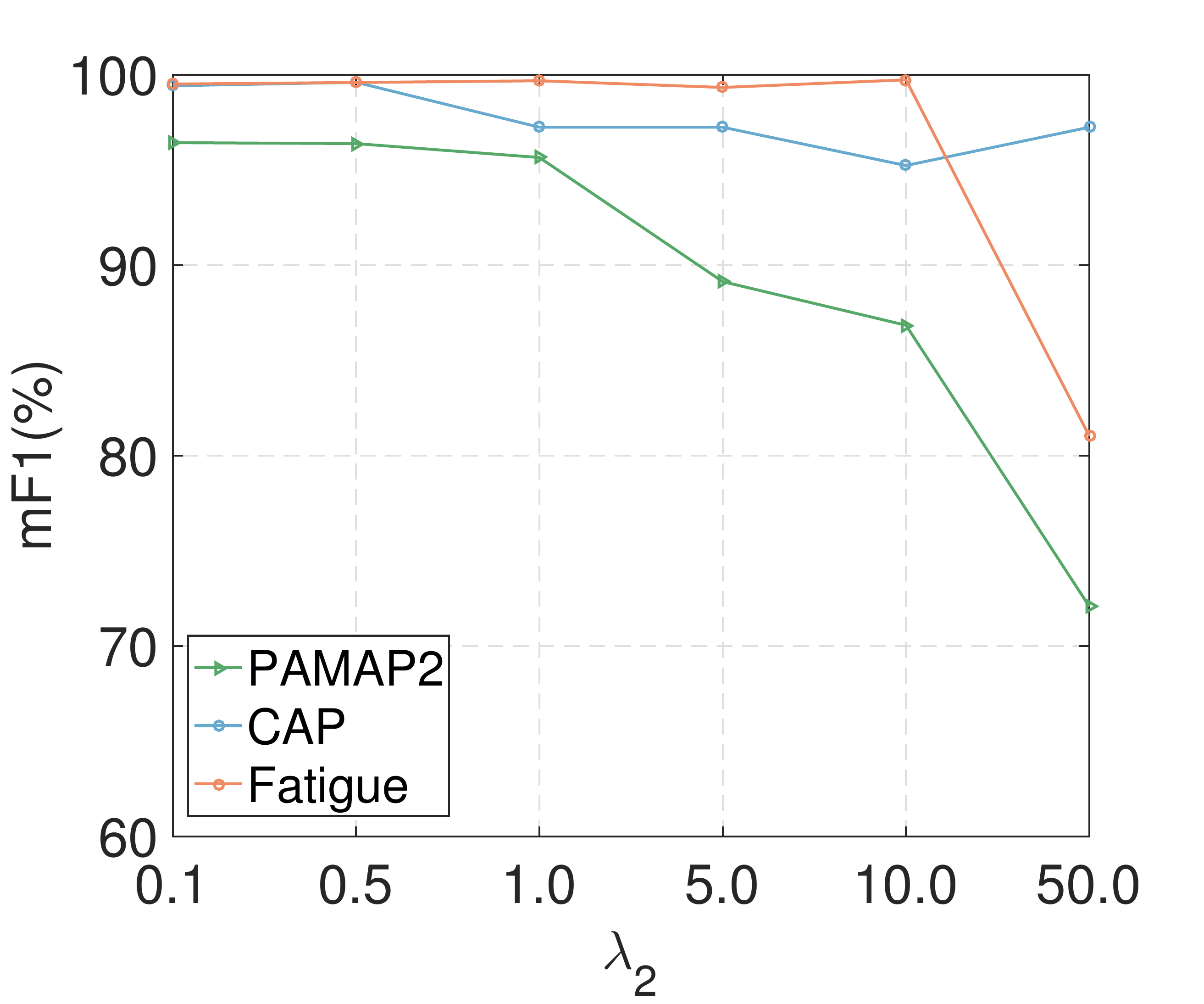}}
	\subfigure[Autoregressive loss]{\includegraphics[width=1.12in]{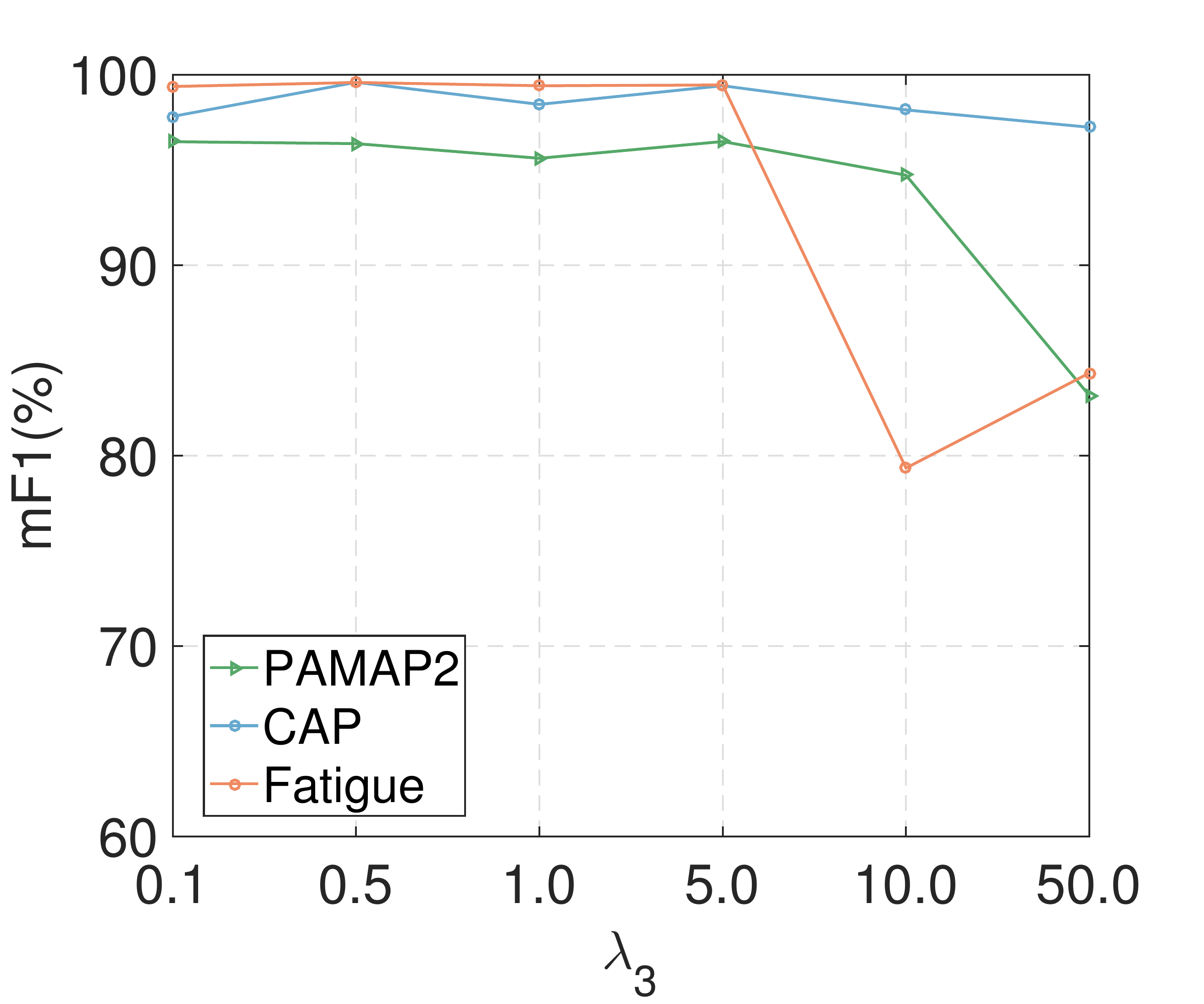}}
	\caption{Parameter sensitivity analysis of the proposed CAE-M approach.}
	\label{fig:sen}
\end{figure}

In this section, we evaluate the parameter sensitivity of CAE-M model. It is worth noting that CAE-M achieves the best performance by adjusting weight coefficient of compound objective function. We apply control variate reduction technique \cite{kucherenko2015application} to empirically evaluate the sensitivity of parameter $\lambda_1,\lambda_2,\lambda_3$ with a wide range. The results are shown in \figurename~\ref{fig:sen}. As the value of MMD loss is greater than others, we select its weight coefficient within e-04 $\sim$ e-07 and other weight coefficients within [0.1, 0.5, 1, 5, 10, 50]. We adjust one of $\lambda$ while fixing the other respective $\lambda$ to keep the optimal value ($\lambda_1= e-04$, $\lambda_2= 0.5$, and $\lambda_3= 0.5$). When weight coefficient is increased, we observe that F1 score tends to decline. The optimal parameter is $\lambda_1= e-04$, $\lambda_2= 0.5$, and $\lambda_3= 0.5$. It can be seen that the performance of CAE-M stays robust within a wide range of parameter choice.

\begin{figure}[htbp]
    \centering
    \subfigure[PAMAP2 dataset]{\includegraphics[width=1.12in]{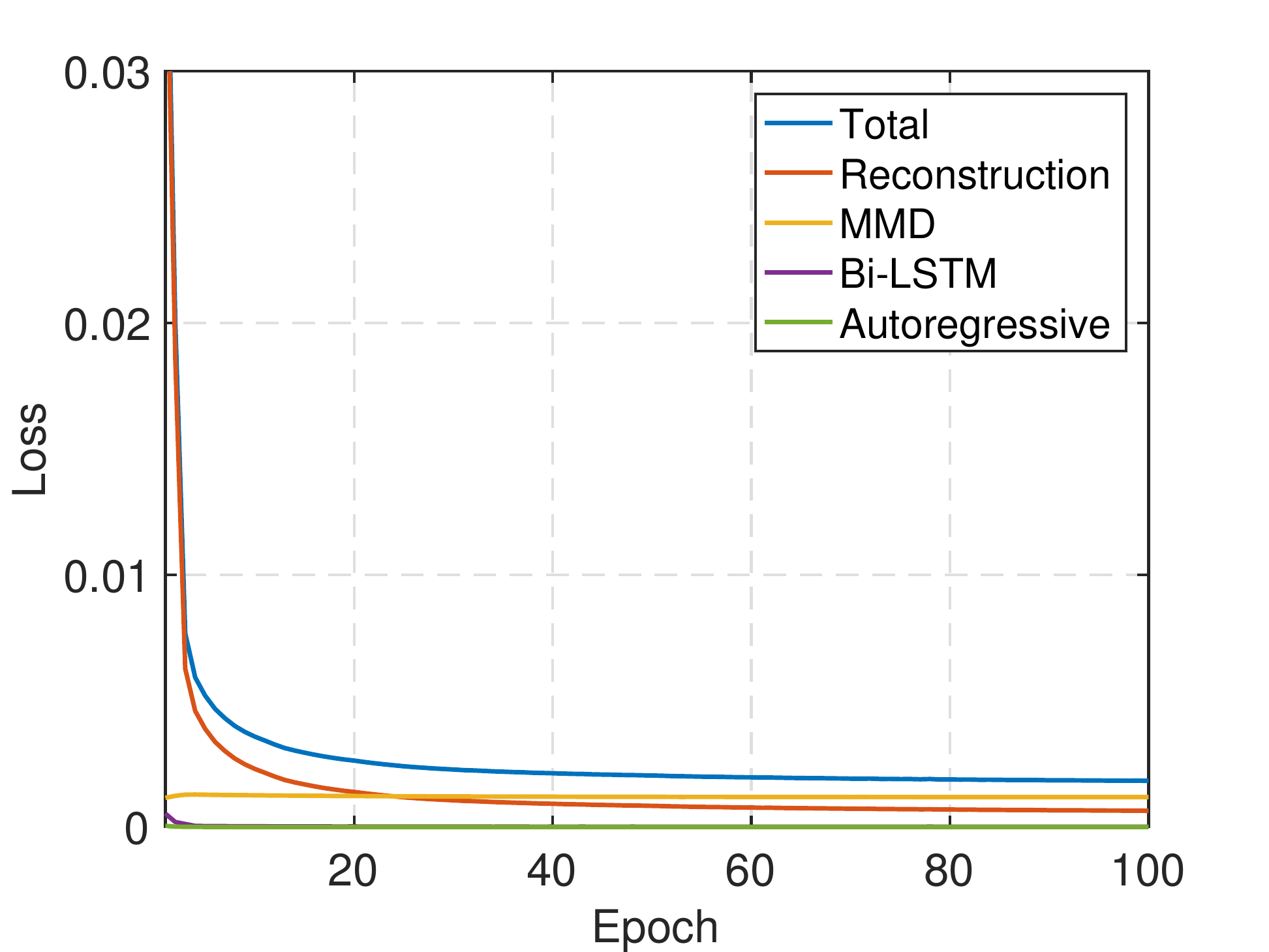}}
    \subfigure[CAP dataset]{\includegraphics[width=1.12in]{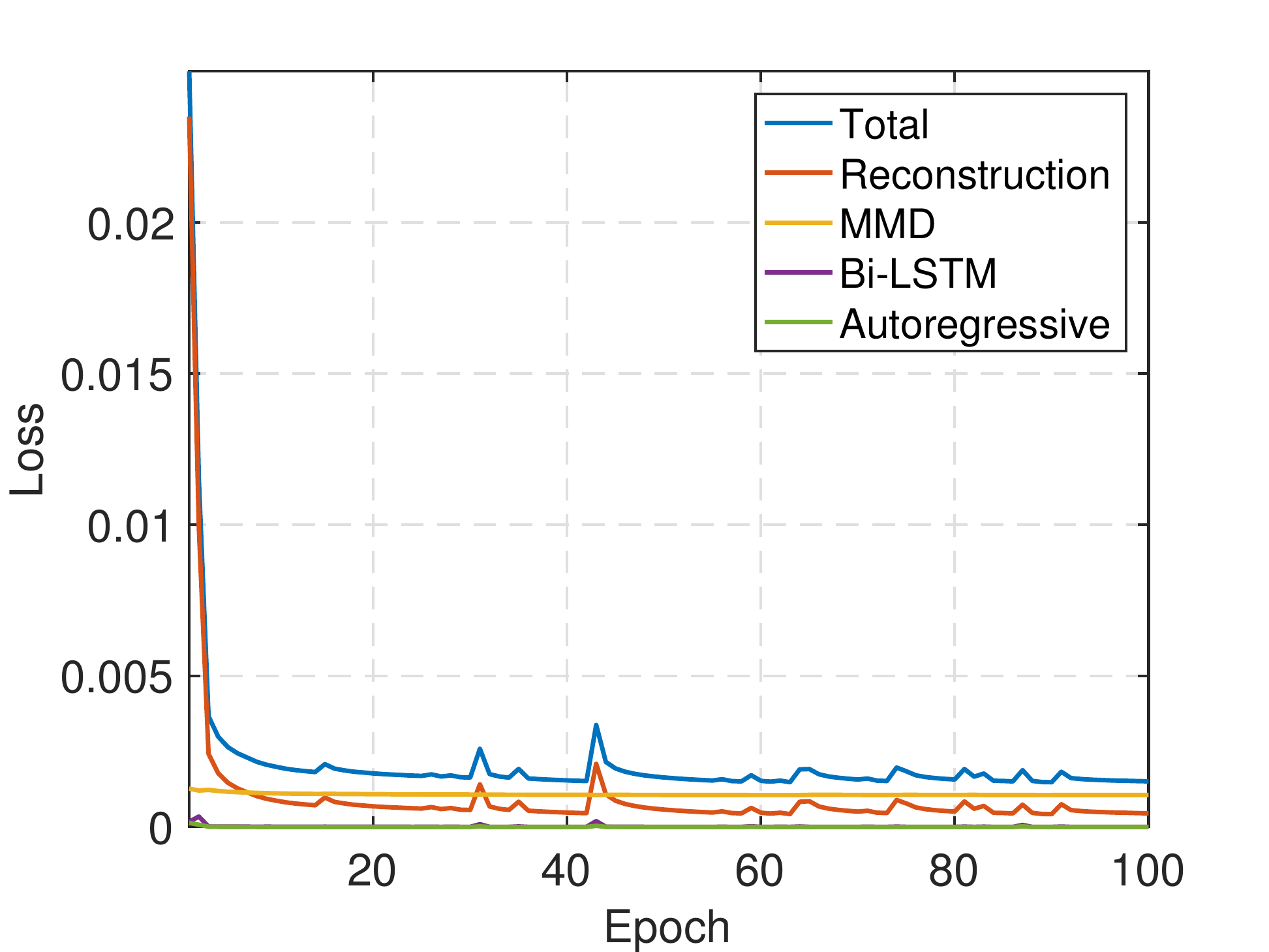}}
    \subfigure[Fatigue dataset]{\includegraphics[width=1.12in]{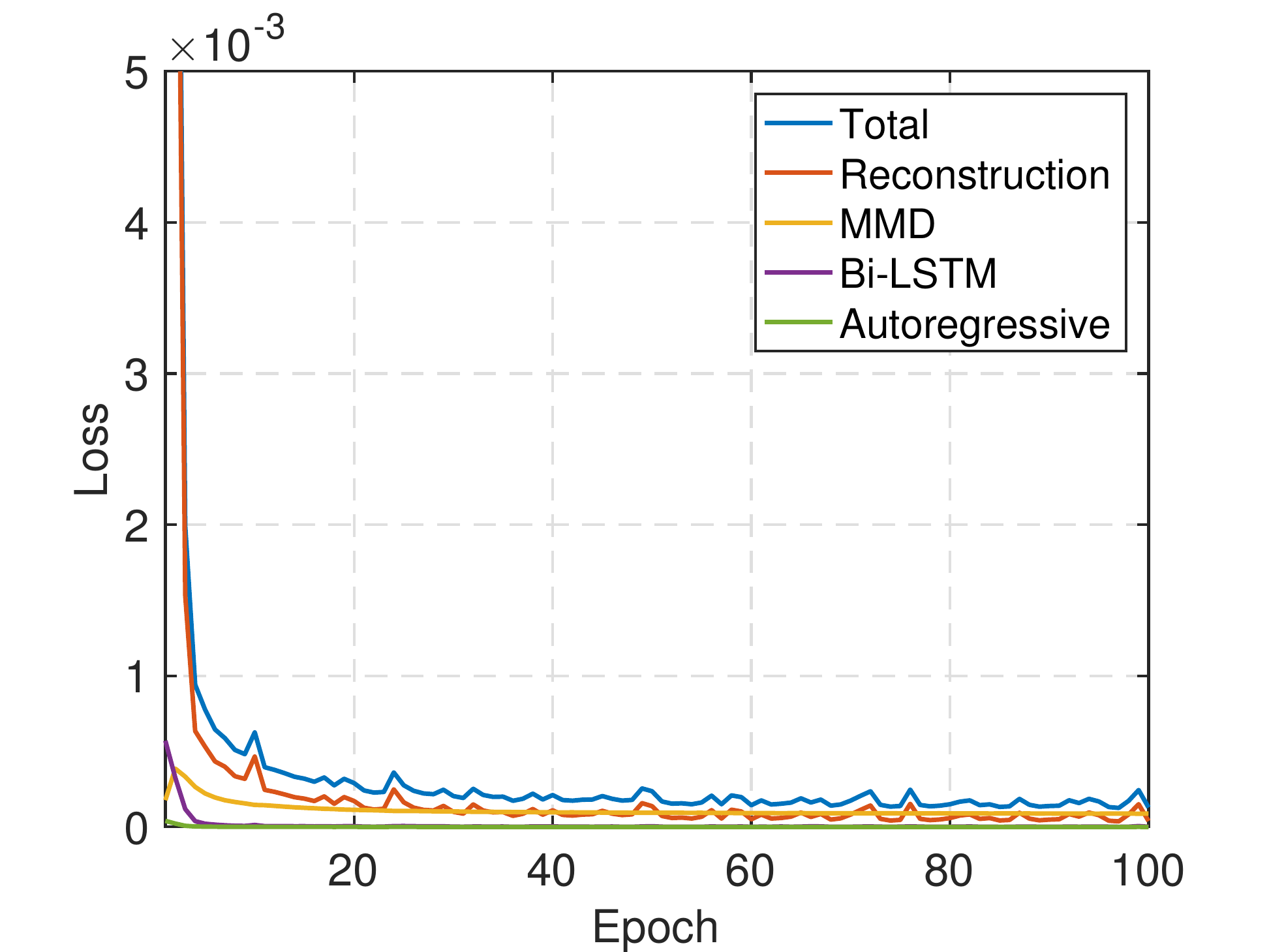}}
    \caption{Convergence analysis of the proposed CAE-M approach on different datasets.}
    \label{fig:con}
\end{figure}

\subsubsection{Convergence Analysis}

Since CAE-M involves several components, it is natural to ask whether and how quickly it can converge. In this section, we analyze the convergence to answer this question. We extensively show the results of each component on three datasets in \figurename~\ref{fig:con}. These results demonstrate that even if the proposed CAE-M approach involves several components, it could reach a steady performance within fewer than 40 iterations. Therefore, in real applications, CAE-M can be applied more easily with a fast and steady convergence performance.

\section{Conclusion and Future Work}
\label{sec:con}
In this paper, we introduced a Deep Convolutional Autoencoding Memory network named CAE-M to detect anomalies. The CAE-M model uses a composite framework to model generalized pattern of normal data by capturing spatial-temporal correlation in multi-sensor time-series data. We first build Deep Convolutional Autoencoder with a Maximum Mean Discrepancy (MMD) penalty to characterize multi-sensor time-series signals and reduce the risk of overfitting caused by noise and anomalies in training data. To better represent temporal dependency of sequential data, we use non-linear Bidirectional LSTM with Attention and linear Auto-regressive model for prediction. Extensive empirical studies on HAR and HC datasets demonstrate that CAE-M performs better than other baseline methods.

In the future work, we will focus on the point-based fine-grained anomaly detection approach and further improve our method for multi-sensor data by designing proper sparse operations.

\section*{Acknowledgment}

This work was supported by Key-Area Research and Development Program of Guangdong Province (No.2019B010109001), Science and Technology Service Network Initiative, Chinese Academy of Sciences (No. KFJ-STS-QYZD-2021-11-001), and Natural Science Foundation of China (No.61972383, No.61902377, No.61902379).

%\appendices
%\section{}

% \ifCLASSOPTIONcompsoc
%   % The Computer Society usually uses the plural form
%   \section*{Acknowledgments}
% \else
%   % regular IEEE prefers the singular form
% \fi

% This work is supported in part by National Key Research \& Development Program of China~(No.2017YFC0803401), CCF-NSFOCUS KunPeng Research Fund~(2018013), and Beijing Municipal Science \& Technology Commission~(No.Z171100000117001).

\ifCLASSOPTIONcaptionsoff
  \newpage
\fi



%\bibliographystyle{IEEEtran}
\bibliographystyle{IEEEtran}
\bibliography{refs}


%\vfill


\end{document}